\documentclass[10pt,journal,compsoc]{IEEEtran}

\ifCLASSOPTIONcompsoc
  \usepackage[nocompress]{cite}
\else
  \usepackage{cite}
\fi

\ifCLASSINFOpdf
\else
\fi

\usepackage[T1]{fontenc}
\usepackage{epsfig}
\usepackage{url}
\usepackage{amsmath,bm}
\usepackage{algorithm}
\usepackage{algorithmic}
\usepackage{subfigure}
\usepackage{epstopdf}
\usepackage{enumerate}
\usepackage{tabularx}
\usepackage{amsthm}
\usepackage{booktabs}
\usepackage{balance}
\usepackage{xspace}
\usepackage{graphicx}
\usepackage{amsfonts}
\usepackage{balance}
\usepackage{color}
\usepackage[table]{xcolor}
\usepackage{makecell}
\usepackage{multirow}
\usepackage{mathtools}
\usepackage{nicematrix_old}

\usepackage[colorlinks=true]{hyperref}

\DeclarePairedDelimiter\floor{\lfloor}{\rfloor}

\theoremstyle{definition}

\newcommand{\method}{\texttt{MTGODE}\xspace}
\definecolor{Red}{RGB}{0,0,0}
\definecolor{Yellow}{RGB}{243,140,10}

\newcommand{\revision}{\textcolor{black}}

\newcommand\algorithmicprocedure{\textbf{def}}
\newcommand{\algorithmicendprocedure}{\algorithmicend\ \algorithmicprocedure}
\makeatletter
\newcommand\PROCEDURE[3][default]{%
  \ALC@it
  \algorithmicprocedure\ \textsc{#2}(#3)%
  \ALC@com{#1}%
  \begin{ALC@prc}%
}
\newcommand\ENDPROCEDURE{%
  \end{ALC@prc}%
  \ifthenelse{\boolean{ALC@noend}}{}{%
    \ALC@it\algorithmicendprocedure
  }%
}
\newenvironment{ALC@prc}{\begin{ALC@g}}{\end{ALC@g}}
\makeatother

\begin{document}

\title{Multivariate Time Series Forecasting with Dynamic Graph Neural ODEs}

\author{Ming Jin, Yu Zheng, 
Yuan-Fang Li, Siheng Chen, Bin Yang, Shirui Pan 
\IEEEcompsocitemizethanks{
\IEEEcompsocthanksitem M. Jin and Y. Li are with the Department of Data Science and AI, Faculty of IT, Monash University, Clayton, Australia \protect 
E-mail: \{ming.jin, yuanfang.li\}@monash.edu;
\IEEEcompsocthanksitem Y. Zheng is with the Department of Computer Science and Information Technology, La Trobe University, Melbourne, Australia \protect 
E-mail: yu.zheng@latrobe.edu.au;
\IEEEcompsocthanksitem S. Chen is with Shanghai Jiao Tong University, Shanghai, China \protect 
E-mail: sihengc@sjtu.edu.cn;
\IEEEcompsocthanksitem B. Yang is with the School of Data Science and Engineering, East China Normal University, China.  \protect 
E-mail: byang@dase.ecnu.edu.cn;
\IEEEcompsocthanksitem S. Pan is is with the School of Information and Communication Technology, Griffith University, Southport, QLD 4222, Australia \protect 
E-mail: shiruipan@ieee.org;
\IEEEcompsocthanksitem Corresponding Author: Shirui Pan
}
\thanks{This research was supported by an ARC Future Fellowship No. FT210100097.}
}

\markboth{Journal of \LaTeX\ Class Files,~Vol.~14, No.~8, August~2015}%
{Shell \MakeLowercase{\textit{et al.}}: Bare Demo of IEEEtran.cls for Computer Society Journals}

\IEEEtitleabstractindextext{%
\begin{abstract}
Multivariate time series forecasting has long received significant attention in real-world applications, such as energy consumption and traffic prediction. While recent methods demonstrate good forecasting abilities, they have three fundamental limitations. (i). \textit{Discrete neural architectures}: Interlacing individually parameterized spatial and temporal blocks to encode rich underlying patterns leads to discontinuous latent state trajectories and higher forecasting numerical errors. (ii). \textit{High complexity}: Discrete approaches complicate models with dedicated designs and redundant parameters, leading to higher computational and memory overheads. (iii). \textit{Reliance on graph priors}: Relying on predefined static graph structures limits their effectiveness and practicability in real-world applications. In this paper, we address all the above limitations by proposing a continuous model to forecast \underline{\textbf{M}}ultivariate \underline{\textbf{T}}ime series with dynamic \underline{\textbf{G}}raph neural \underline{\textbf{O}}rdinary \underline{\textbf{D}}ifferential \underline{\textbf{E}}quations (\method). Specifically, we first abstract multivariate time series into dynamic graphs with time-evolving node features and unknown graph structures. Then, we design and solve a neural ODE to complement missing graph topologies and unify both spatial and temporal message passing, allowing deeper graph propagation and fine-grained temporal information aggregation to characterize stable and precise latent spatial-temporal dynamics. Our experiments demonstrate the superiorities of \method from various perspectives on five time series benchmark datasets.
\end{abstract}

\begin{IEEEkeywords}
multivariate time series forecasting, graph neural networks, neural ordinary differential equations.
\end{IEEEkeywords}}

\maketitle
\IEEEdisplaynontitleabstractindextext
\IEEEpeerreviewmaketitle
\IEEEraisesectionheading{\section{Introduction}\label{sec:introduction}}

\IEEEPARstart{T}{ime} series data plays a vital role in shaping modern societies and has long been studied across multiple fields in science and engineering, such as energy grid balancing\cite{heidrich2020forecasting}, climate studies\cite{li2021weather}, and traffic volume forecasting\cite{yu2018spatio}. Among these applications and given a sensor network, the multivariate time series data can be interpreted as the combination of recorded univariate time series on each sensor, which can be interconnected and mutually influenced. For example, the rise in daily average temperature may lead to an increase in traffic volume on coastal roads. Therefore, multivariate time series forecasting largely depends on modeling the underlying spatial-temporal correlations, which directly affects the reliability of those above and many other real-world applications.

Nevertheless, this task is not easy because it is principally challenging to effectively and efficiently model the underlying complex spatial-temporal dependencies on multivariate time series. While earlier methods are based on statistical models \cite{lutkepohl2013vector, zhang2003time, frigola2015bayesian}, recent works take deep learning-based approaches, demonstrating better capabilities to capture nonlinear temporal and spatial patterns. Although recurrent neural networks (RNNs) \cite{rumelhart1986learning} have been widely adopted in aggregating temporal information on time series data, they suffer from certain limitations when processing long sequences, such as time-consuming iteration and gradient explosion \revision{\cite{wu2019graph}}. In modeling multivariate time series, vanilla RNNs and their variants \cite{chung2014empirical, hochreiter1997long, schuster1997bidirectional} also fail to exploit the dynamic interdependencies among variables. To address the above limitations, LSTNet \cite{lai2018modeling} adopts 1D convolution neural network (CNN) and two RNN variants to capture short-term local variable dependencies and long-term temporal patterns. TPA-LSTM \cite{shih2019temporal} first processes the input sequences via an RNN and then leverages multiple 1D convolution filters and a scoring function to capture both temporal and spatial correlations. HyDCNN \cite{li2021modeling}, on the other hand, designs a CNN-based model to capture rich spatial and temporal patterns simultaneously. However, these methods do not explicitly model the pairwise dependencies between variables, limiting their effectiveness in forecasting multivariate time series.

Recently, techniques based on graph neural networks (GNNs) \cite{jiang2021graph} have demonstrated great potential in modeling the spatial and temporal interdependencies simultaneously among multiple time series over time. As a specific data format, graphs can naturally be adopted to describe the interconnections between entities. In the context of multivariate time series forecasting, the essence of these methods is predicting future node features with the help of historical observations and predefined graph structures, where nodes, node features, and static edges are variables, univariate time series, and the prior knowledge to describe stable relationships between variables (e.g., metro networks), respectively.

\begin{figure}[t]
    \centering
    \includegraphics[width=7cm]{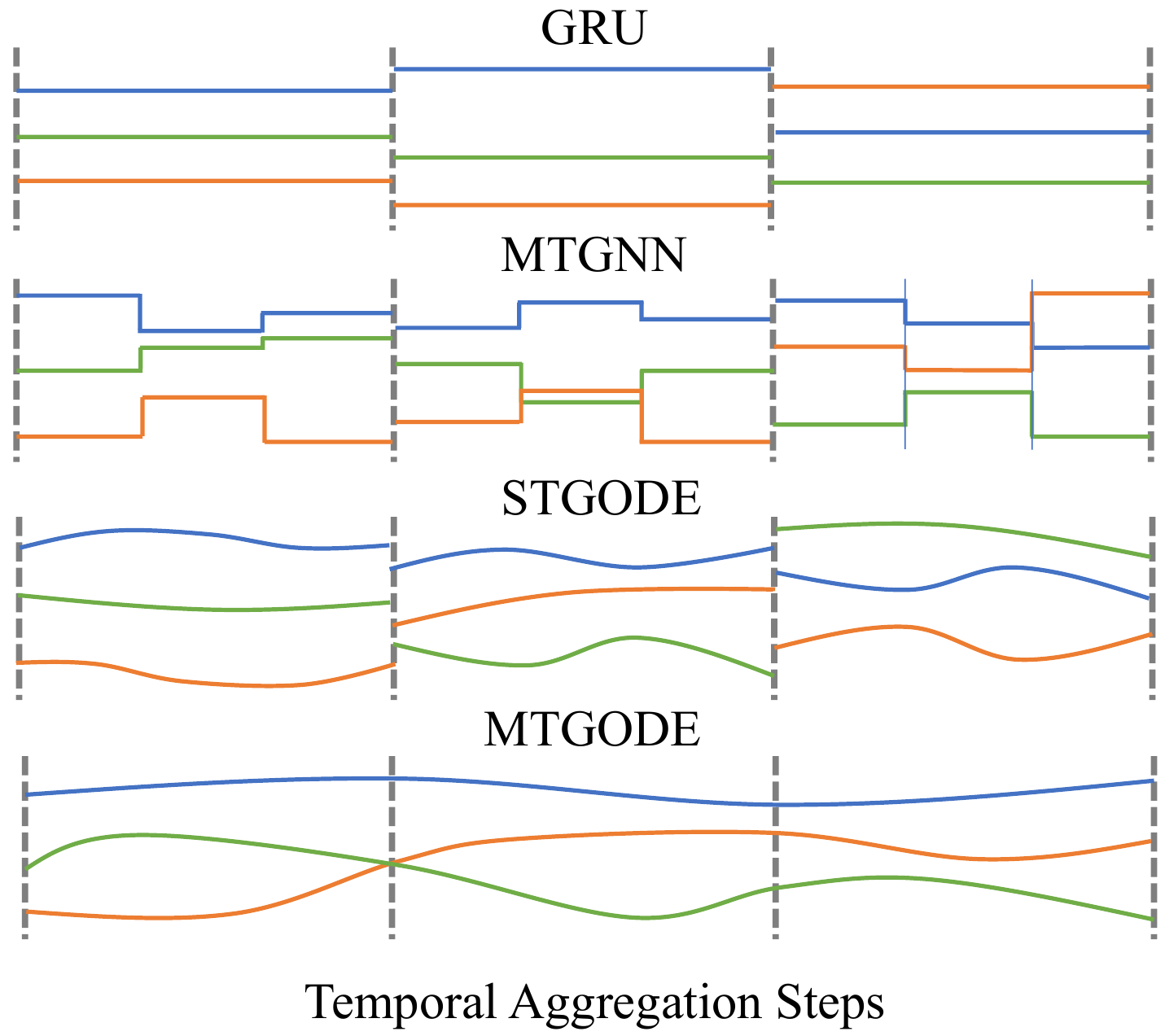}
    \vspace{-1mm}
    \caption{The latent state trajectories of different forecasting methods when encoding historical observations. The vertical dash lines denote temporal aggregation steps, and the three colored solid lines are latent state trajectories of different variables. Among these methods, GRU has invariant latent states between two observations, while MTGNN has a sequence of transformations (i.e., graph propagation) after each temporal aggregation step. Although STGODE generates the discrete graph propagation with an ODE, its temporal aggregation remains discrete. In contrast, \method allows modeling the fully continuous latent spatial-temporal dynamics.}
    \vspace{-3mm}
    \label{fig:latent trajectories}
\end{figure}

For instance, DCRNN \cite{li2018diffusion} proposes a bidirectional graph random walks-based gated recurrent unit to model spatial and temporal dependencies, STGCN \cite{yu2018spatio} intersects graph and temporal convolutions to learn on multivariate time series data, and GMAN \cite{zheng2020gman} designs a spatial-temporal block to do similar things by composing two attention mechanisms on the graph and temporal spaces. Although these methods demonstrate competitive performances, it remains difficult for them to accurately model arbitrary multivariate time series based on the following challenges:

\begin{itemize}
    \item \textit{Challenge 1: Discrete Neural Architectures.} 
    Instead of parameterizing the continuous dynamics of latent states, the existing works on modeling multivariate time series are based on entirely or partially discrete neural architectures, \revision{resulting in discontinuous state trajectories in modeling latent spatial-temporal dynamics, which is shown to be less effective \cite{chen2018neural} and thus hinders downstream tasks in terms of the forecasting precision.}
    \revision{Figure \ref{fig:latent trajectories} plots the latent state trajectories of four typical methods when learning on multivariate time series data. Specifically, most of these approaches fail to define a vector field to characterize the fully continuous latent spatial-temporal dynamics except for our method, which demonstrates significantly better downstream forecasting performance (Tables \ref{table: single-step forecasting results} and \ref{table: multi-step forecasting results}).}
    Another limitation of discrete neural architectures is the shallow graph propagation in most GNN-based approaches, such as STGCN \cite{yu2018spatio} and MTGNN \cite{wu2020connecting}, due to the challenge of over-smoothing, which prevents them from considering the spatial correlations from farther neighbors and further limiting their forecasting abilities.
    
    \item \textit{Challenge 2: High Complexity.} 
    Discretely stacking individually parameterized spatial and temporal modules, e.g., in \cite{yu2018spatio} and \cite{wu2020connecting}, not only results in discontinuous latent state trajectories but also complicates models with dedicated designs (e.g., \revision{parameterized} residual and skip connections) and redundant trainable parameters, leading to computational and memory inefficient. We theoretically and empirically justify this in Subsections \ref{subsec:model training}, \ref{subsec:study on two ODEs}, and \ref{subsec:computational efficiency}.
    
    \item \textit{Challenge 3: Rely on Graph Priors.} Plenty of existing GNN-based forecasting models, e.g., \cite{yu2018spatio}, \cite{zheng2020gman}, \cite{chen2020multi}, and \cite{fang2021spatial}, require prior knowledge of graph structures (i.e., stable interconnections between variables). However, such knowledge is typically unknown in most cases, hindering their applications in broader real-world applications.
\end{itemize}

\revision{Although some recent works aim to address the above limitations, none of these methods can solve them all.}
\revision{For example, STGODE \cite{fang2021spatial} proposes an ODE network to characterize the continuous propagation on predefined graphs, but its temporal aggregation process remains discrete. Thus, it faces all the above challenges.}
\revision{Other methods, such as GTS \cite{shang2020discrete} and MTGNN \cite{wu2020connecting}, get rid of the predefined graph structures, but they leave the first and second limitations unsolved.}
\revision{While a recently proposed method, STG-NCDE \cite{choi2022STGNCDE}, addresses the first and third challenges with Neural Controlled Differential Equations (NCDEs) \cite{kidger2020neural} and graph structure learning, it remains complex and less effective when modeling long input series because of its recursive nature and interpolating preprocessing.}

In this paper, we resolve all the above challenges by proposing a novel method to forecast \underline{\textbf{M}}ultivariate \underline{\textbf{T}}ime series with dynamic \underline{\textbf{G}}raph neural \underline{\textbf{O}}rdinary \underline{\textbf{D}}ifferential \underline{\textbf{E}}quations (\method\footnote{\revision{Code is available at \url{https://github.com/GRAND-Lab/MTGODE}}} for abbreviation).
Compared with existing works, our theme allows explicitly encoding the fully continuous spatial-temporal dynamics of arbitrary multivariate time series in the latent space, which benefits various downstream forecasting tasks by a large margin.
Specifically, we abstract input series as dynamic graphs with time-evolving node features and unknown graph structures. 
To complement and learn from the missing interdependencies between nodes (i.e., variables), we propose a \textit{continuous graph propagation} mechanism together with a graph structure learning schema to partially and wholly address the first and third challenges, which significantly alleviates the over-smoothing issue in GNNs and thus allows deeper continuous propagation on dynamically self-extracted graph structures to capture long-range spatial correlations between time series.
\revision{To encode rich temporal information and completely resolve the first challenge, we propose a \textit{continuous temporal aggregation} mechanism to parameterize the derivative of latent states instead of themselves, allowing fine-grained temporal patterns to be extracted and aggregated precisely.  It is worth noting that this mechanism also tactfully addresses the second challenge by eliminating redundant computations and disentangling the ties between aggregation depth and memory bottleneck in discrete formulations, thus can provide more accurate modeling of latent temporal dynamics than discrete methods with limited computational budgets.}
In \method, we elegantly couple two mechanisms and provide a simpler framework than most GNN-based forecasting pioneers to learn from and forecast multivariate time series that are both more effective and efficient, thus resolving the three aforementioned challenges. We summarize our contributions as follows:
\begin{itemize}
    \item To the best of our knowledge, this is the first work to learn fully continuous \revision{latent spatial-temporal dynamics of arbitrary multivariate time series} by unifying spatial and temporal message passing with two \revision{coupled ODEs} and a more concise model design.
    \item We propose a \revision{spatial ODE} together with a graph learning schema to learn continuous long-range spatial dynamics between time series, which alleviates the reliance on static graph priors and the common over-smoothing problem in GNNs.
    \item We propose a temporal neural ODE by generalizing canonical temporal convolutions to learn the continuous fine-grained temporal dynamics of time series, \revision{resulting in a powerful and efficient forecasting model with the proposed spatial ODE.}
    \item We conduct extensive experiments to demonstrate the effectiveness and efficiency of the proposed method, showing better application prospects.
\end{itemize}

We organize the rest of the paper as follows: Section \ref{sec:related work} reviews the related work. Section \ref{sec:problem definition} provides the problem definition and notations. Section \ref{sec:methodology} presents the proposed method and algorithms. In Section \ref{sec:experiment} and \ref{sec:conclusion}, we discuss the experimental results and conclude the paper.

\section{Related Work} \label{sec:related work}
This work is closely related to multivariate time series forecasting, graph neural networks, and neural ordinary differential equations. We briefly review related representative works in this section.

\subsection{Multivariate Time Series Forecasting}
Multivariate time series forecasting has long been a widely studied subject, where existing approaches are in two categories: Statistical and deep learning-based. 
For the former branch of methods, auto-regressive (AR) models linearly predict future changes in a time series based on historical observations. Vector auto-regressive (VAR) approaches \cite{lutkepohl2013vector} further extend AR by exploring the interdependencies between multiple time series. ARIMA \cite{box2015time}, on the other hand, integrates the ideas of AR and moving average (MA). On top of this, VARIMA \cite{de200625} generates ARIMA to operate on multivariate time series. Although statistical models are widely applied in real-world applications because of their interpretability and simplicity, they only explore linear relationships and make strong assumptions of stationary processes. In contrast, recent deep learning-based methods are free from these limitations and demonstrate better performances. 
LSTNet \cite{lai2018modeling} and TPA-LSTM \cite{shih2019temporal} are two models that propose to learn discrete temporal dynamics and local spatial correlations between time series via RNNs and CNNs. Recently, to address the parallelization issue in RNNs, methods built on CNNs or Transformer \cite{vaswani2017attention} demonstrate a better efficiency and forecasting ability. For example, HyDCNN \cite{li2021modeling} adopts position-aware dilated CNNs to model both spatial and temporal information, and Informer \cite{zhou2021informer} does the same things with a proposed variant of vanilla Transformer. However, all those methods have not explicitly modeled the pairwise dependencies between variables, limiting their effectiveness in forecasting multivariate time series. 
In this work, we first abstract multivariate time series as dynamic graphs with time-evolving node features and unknown graph structures, and then complement and learn from the missing interdependencies between nodes with the proposed continuous graph propagation mechanism, which allows \method better to capture the pairwise dependencies between nodes (i.e., variables) explicitly in arbitrary multivariate time series.

\begin{table}[t]
	\small
	\centering
	\caption{Summary of the primary notations.} 
	\begin{tabular}{ p{77 pt}<{\centering} | p{155 pt}}  
		\toprule[1.0pt]
		Symbols & Description  \\
		\cmidrule{1-2}
		$\mathbf{X} \in \mathbb{R}^{N \times D \times S}$ & A given multivariate time series data \\
		$\mathbf{X}_{t+1 : t+T} \in \mathbb{R}^{N \times D \times T}$ & A sequence of $T$ historical observations sampled from $\mathbf{X}$ \\
		$\mathbf{A} \in \mathbb{R}^{N \times N}$ & \revision{A learned adjacency matrix} \\
		\cmidrule{1-2}
        $\mathbf{H}^{G}_{k} \in \mathbb{R}^{N \times D' \times Q}$ & The latent state of discrete graph propagation at $k$-th layer \\
        $\mathbf{H}^{G}(t) \in \mathbb{R}^{N \times D' \times Q}$ & The intermediate latent state of CGP process at $t$ \\
        $\mathbf{H}^{T}_{l} \in \mathbb{R}^{N \times D' \times Q_l}$ & The latent state of discrete temporal aggregation at $l$-th layer \\
        $\mathbf{H}^{T}(t) \in \mathbb{R}^{N \times D' \times R}$ & The intermediate latent state of CTA process at $t$ \\
        $\mathbf{H}^{G}_{out} \in \mathbb{R}^{N \times D' \times Q}$ & The learned spatial representation of the graph module \\
        $\mathbf{H}^{T}_{out} \in \mathbb{R}^{N \times D'}$ & The learned temporal representation of the temporal module \\
        $\mathbf{H}_{out} \in \mathbb{R}^{N \times D'}$ & The learned spatial-temporal representation of $\mathbf{X}_{t+1 : t+T}$ \\
        $\mathbf{\Theta}, \mathbf{\Phi}, \mathbf{\Gamma}$ & The trainable parameters of \method \\
		\cmidrule{1-2}
		$N$, $S$, $D$ & The number of variables, length, and feature \revision{dimensions} of $\mathbf{X}$ \\
        \revision{$T$, $H$} & \revision{Input length and forecasting horizon} \\
		$D'$ & The output \revision{dimensions} of $\mathbf{H}_{out}$ \\
        \revision{$K$, $L$} & \revision{The number of layers of discrete graph propagation and temporal aggregation} \\
		$T_{cgp}, \Delta t_{cgp}$ & The integration time and \revision{step size when defining a CGP process} \\
		$T_{cta}, \Delta t_{cta}$ & The integration time and \revision{step size when defining a CTA process} \\
		\bottomrule[1.0pt]
	\end{tabular}
	\label{table:notation}
\end{table}

\subsection{Graph Neural Networks}
\revision{Graphs are ubiquitous in the real world, and GNNs are designed to incorporate attributive and topological information to learn expressive node-level or graph-level representations \cite{zhang2022trustworthy, liu2022graph}, where spatial correlations between nodes are explicitly modeled by passing messages from nodes' neighbors to nodes themselves.}
Recently, several works have emerged to tackle the traffic forecasting problem with GNN-based models \cite{li2018diffusion, yu2018spatio, wu2019graph, zheng2020gman, chen2020multi, wu2020connecting, shang2020discrete, fang2021spatial, choi2022STGNCDE}. 
Given an input multivariate time series and a predefined graph structure to characterize the static relationships between variables (i.e., nodes), they typically adopt graph convolutions to capture local spatial dependencies and use RNNs \cite{li2018diffusion, shang2020discrete}, or 1D convolutions \cite{yu2018spatio, wu2019graph, wu2020connecting} to model temporal dynamics. Although minor works exist to alleviate the reliance on graph priors \cite{shang2020discrete, wu2020connecting, choi2022STGNCDE} or conduct deeper graph propagation \cite{fang2021spatial} to capture long-range spatial dependencies, \revision{they fail to completely address all three above challenges to effectively and efficiently learn stable and precise spatial-temporal dynamics on arbitrary multivariate time series data in the latent space.}
\revision{To bridge the gaps}, we propose a simpler model by elegantly coupling two proposed continuous mechanisms, demonstrating significantly better effectiveness and efficiency.

\subsection{Neural Ordinary Differential Equations}
Chen \textit{et al.} \cite{chen2018neural} introduced a new paradigm of continuous-time models by generalizing discrete deep neural networks. Taking a $L$-layer residual network as an example, it can be formulated as follows:
\begin{equation}
\left\{
\begin{aligned}
& \mathbf{H}_{l+1} = \mathbf{H}_l + f(\mathbf{H}_{l}, \mathbf{\Theta}_l), \\
& \mathbf{H}_{out} = \mathbf{H}_L.
\end{aligned}
\right.
\label{eq: residual networks}
\end{equation}
If we insert more layers and take smaller integration steps, then we can directly parameterize and approximate the continuous evolution of latent states, which forms the basic idea of Neural Ordinary Differential Equations (NODEs):
\begin{equation}
\left\{
{\begin{aligned}
&\frac{\displaystyle\mathrm{d}\mathbf{H}(t)}{\displaystyle\mathrm{d}t} = f(\mathbf{H}(t), \mathbf{\Theta}), \\
&\mathbf{H}_{out} = \operatorname{ODESolve}\big(\mathbf{H}(0), f, t_0, t_1, \mathbf{\Theta}\big).
\end{aligned}} \right.
\label{eq: neural ODE}
\end{equation}
\revision{In the following, we omit $t_0$ in the above equation for simplicity if $t_0=0$.}
Recently, NODEs have been adopted in some research fields, \revision{such as graph neural networks \cite{xhonneux2020continuous, wang2021dissecting} and traffic forecasting \cite{fang2021spatial}.} 
Specifically, as the only ODE-based method for traffic forecasting, STGODE \cite{fang2021spatial} merely considers the continuous graph propagation on predefined static graph structures without modeling the continuous temporal dynamics. Our approach distinguishes from it in two important aspects. Firstly, we propose a novel continuous temporal aggregation mechanism coupled with a simplified continuous graph propagation process to learn more expressive latent spatial-temporal dynamics efficiently in a fully continuous manner. Secondly, our method eliminates the reliance on predefined graph structures.
\revision{On the other hand, as an extension of NODEs, Neural Controlled Differential Equations (NCDEs) \cite{kidger2020neural} emerges as a continuous generalization of RNNs to learn on time series data naturally. A recently proposed method, STG-NCDE \cite{choi2022STGNCDE}, further extends this idea to model traffic data with two different NCDEs to severally model temporal and spatial dependencies, showing good forecasting results. Although STG-NCDE learns continuous latent dynamics without relying on predefined graph priors, it cannot effectively and efficiently handle long input series due to its recursive nature like in RNNs and resource-intensive interpolating preprocessing. In contrast, our method is free from this issue with a non-recursive backbone process and a light preprocessing module.}
\revision{Thus, compared with the above two methods, \method can efficiently forecast arbitrary multivariate time series with more competitive performance.}

\section{Problem Definition} \label{sec:problem definition}
In this section, we introduce the problem of representation learning on multivariate time series with two commonly adopted evaluation protocols, i.e., single-step and multi-step forecasting.
Specifically, the bold uppercase and lowercase letters denote matrices and vectors.
We summarize all important notations in Table \ref{table:notation}.

\begin{figure*}[t]
    \centering
    \includegraphics[width=0.97\textwidth]{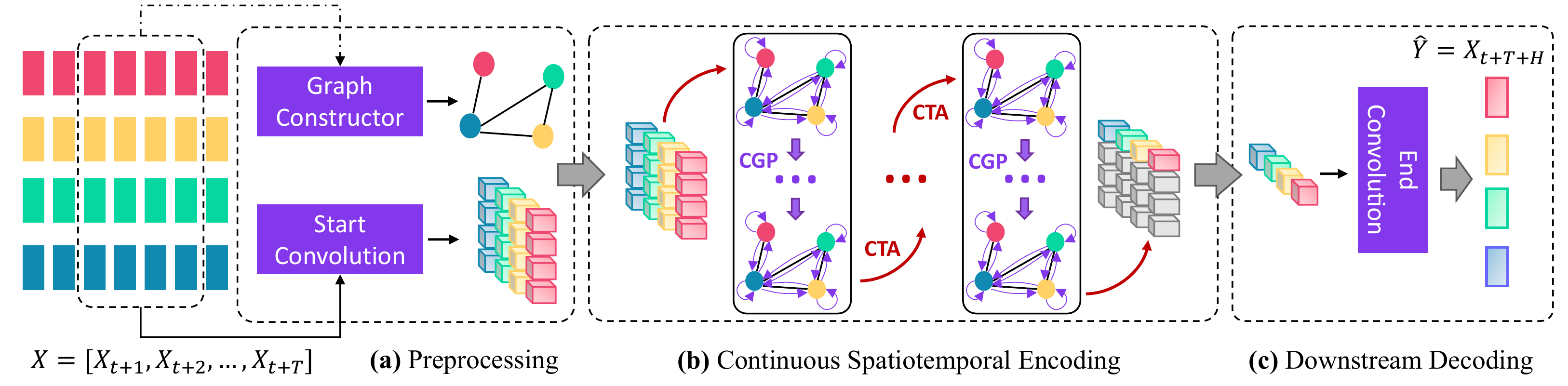}
    \caption{The overall framework of \method. Given a sequence of historical observations, we first map them to the latent space and learn an associated graph structure. Then, the continuous dynamics of spatial and temporal clues are modeled by coupling two ODEs from different perspectives, namely the \textit{continuous temporal aggregation} (CTA) and \textit{continuous graph propagation} (CGP) in the middle part. Finally, the learned representations can be used in various forecasting tasks, such as the plotted single-step forecasting in the rightmost part.}
    \label{fig:framework}
\end{figure*}

Let $\mathbf{X} \in \mathbb{R}^{N \times D \times S}$ denote a multivariate time series with $N$ variables, $D$ feature dimensions, and $S$ time steps in total for training. Specifically, we define $\mathbf{X}^i \in \mathbb{R}^{D \times S}$ as the $i$-th time series for all features and time steps, and $\mathbf{X}_{t} \in \mathbb{R}^{N \times D}$ as the $t$-th time step for all series and features.
Given a sequence of $T$ historical observations $\mathbf{X}_{t+1 : t+T} \in \mathbb{R}^{N \times D \times T}$, our objective is to learn a spatial-temporal encoder $f(\cdot): \mathbb{R}^{N \times D \times T} \rightarrow \mathbb{R}^{N \times D'}$, where the learned representation $\mathbf{H}_{out}=f(\mathbf{X}_{t+1 : t+T})$ can be used in various downstream tasks, such as the single-step and multi-step forecasting of future observations in Section \ref{sec:experiment}. Formally, given a loss function $\ell(\cdot)$ and for each valid time step $t$, we formulate the problem of multivariate time series forecasting as follows:
\begin{equation}
    f^*, g^* = \mathop{\arg\min}\limits_{f, g}\sum_t\ell\Big(g\big(f(\mathbf{X}_{t+1 : t+T})\big), \mathbf{Y}\Big),
    \label{eq: problem}
\end{equation}
where $f^*(\cdot)$ and $g^*(\cdot)$ represent the encoder and decoder with learned optimal parameters.
Specifically, we let $\mathbf{Y}=\mathbf{X}_{t+T+H} \in \mathbb{R}^{N \times D \times 1}$ for single-step forecasting, and $\mathbf{Y}=\mathbf{X}_{t+T+1 : t+T+H} \in \mathbb{R}^{N \times D \times H}$ for multi-step forecasting. $H$ represents a specific forecasting horizon.

\section{Methodology} \label{sec:methodology}
\revision{In this section, we present the overall framework and detailed designs of the proposed \method method.}
As shown in Figure \ref{fig:framework}, our method mainly consists of three main components, namely the \textit{data preprocessing}, \textit{continuous spatial-temporal encoding}, and \textit{downstream task decoding}.
Given a sequence of multivariate time series, we first map it to the latent space with a $1 \times 1$ convolution kernel and acquire the underlying topological structure dynamically with a graph constructor.
\revision{Then, to extract and encode rich interlaced spatial-temporal patterns of input series, we propose two elegantly coupled continuous processes, i.e., \textit{continuous graph propagation (CGP)} and \textit{continuous temporal aggregation (CTA)}}
Different from \cite{yu2018spatio}, \cite{wu2020connecting}, and \cite{fang2021spatial}, our former process with the graph structure learning not only enables the model to complement and learn from the missing interdependencies between time series but also allows the continuous and farther message passing on dynamically self-extracted graph structures to learn continuous long-range spatial dynamics. 
\revision{To effectively extract and aggregate fine-grained temporal patterns, we surrogate discrete temporal convolutions with the proposed CTA process, which defines a vector field to drive and model the underlying latent temporal dynamics continuously.}
\revision{Specifically, we parameterize the derivative of latent states instead of themselves, providing more accurate modeling of complex temporal dependencies, eliminating redundant computations (e.g., parameterized residual and skip connections), and disentangling the ties between aggregation depth and memory bottleneck in discrete formulations, thus resulting in better model effectiveness and efficiency.}
\revision{In \method, the proposed continuous spatial and temporal message passing mechanisms are elegantly unified; thus, our approach can effectively and efficiently model complex latent spatial-temporal dynamics of arbitrary multivariate time series in a fully continuous manner.}
Finally, given the learned representations of historical observations, we can conduct various forecasting tasks by employing different downstream decoders, e.g., the single-step forecasting in the rightmost part \revision{of} Figure \ref{fig:framework}.

In the rest of this section, \revision{we introduce the proposed CGP and CTA processes in Subsection \ref{subsec:CGP} and \ref{subsec:CTA}.}
In Subsection \ref{subsec:theoritical comparision}, we theoretically analyse \revision{our method} by comparing with their discrete variants. The details of model training and algorithms are discussed in Subsection \ref{subsec:model training}.

\subsection{Continuous Graph Propagation}\label{subsec:CGP}
In \method, we essentially integrate temporal aggregation and graph propagation processes to capture rich spatial-temporal patterns from historical observations.
\revision{Specifically, at each temporal aggregation step and for simplicity, the spatial dependencies between time series can be characterized by the combination of feature propagation and linear transformation on a specific graph snapshot with the feature matrix $\mathbf{H}^{G}_{0} \in \mathbb{R}^{N \times D' \times Q}$, which is the latent states at a specific aggregation step (see Subsections \ref{subsec:CTA} and \ref{subsec:model training} for details).} 
\revision{In a nutshell, given an adjacency matrix $\mathbf{A} \in \mathbb{R}^{N \times N}$ and initial states $\mathbf{H}^{G}_{0}$,} a discrete formulation of the $K$-hop graph propagation is defined as \cite{wu2019simplifying}:
\begin{equation}
\left\{
\begin{aligned}
        & \mathbf{H}^{G}_{k+1} = \widehat{\mathbf{A}} \ \mathbf{H}^{G}_{k},\  k \in \{0, \cdots, \revision{K-1}\}, \\  
        & \mathbf{H}_{out}^{G} = \mathbf{H}^{G}_{K} \ \mathbf{\Phi},
\end{aligned}
\right.
\label{eq: SGC}
\end{equation}
where $\widehat{\mathbf{A}}$ denotes the normalized adjacency matrix, $\mathbf{H}_{out}^{G} \in \mathbb{R}^{N \times D' \times Q}$ is the output \revision{representations}, and $\mathbf{\Phi} \in \mathbb{R}^{D' \times D'}$ is a trainable parameter matrix. In practice, we define tensor multiplication in feature propagation with the Einstein summation in the above equation to sum the element products along specific dimensions.
This is because the feature propagation only operates on the first two dimensions of latent states without aggregating information along the time axis (with the sequence length $Q$).

Compared with GCN \cite{kipf2017semi}, Equation \ref{eq: SGC} eliminates the redundant nonlinearities and further decouples the feature propagation and transformation steps, resulting in a simpler and more efficient model while maintaining comparable accuracy. However, this discrete formulation is error-prone and vulnerable to over-smoothing when conducting a deep propagation on graphs. 
The underlying cause of these two problems in Equation {\protect\ref{eq: SGC}} can be uncovered by decomposing the propagation depth $K$ into the combination of integration time $T_{cgp}$ and \revision{step size} $\Delta t_{cgp}$, i.e., $K=T_{cgp} / \Delta t_{cgp}$. From the perspective of a continuous process, a selected $T_{cgp}$ and $ \Delta t_{cgp}$ control the number of function evaluations, which is equivalent to describing how many times feature propagation is executed, a.k.a. the propagation depth $K$ in the discrete formulation.
Therefore, considering a case where a fixed integration time and smaller \revision{step size} are applied, we can naturally have the following transformation with propagation steps $k \in \{0, \cdots, \revision{K-1}\}$ being replaced by a continuous variable $t \in \mathbb{R}_{0}^{+}$:
\begin{equation}
\begin{aligned}
        \mathbf{H}^{G}(t+\Delta t_{cgp}) &= \mathbf{H}^G(t) + \Delta t_{cgp} (\widehat{\mathbf{A}}-\mathbf{I}_N)\mathbf{H}^G(t) \\  
        &= \big[(1-\Delta t_{cgp})\mathbf{I}_N + \Delta t_{cgp} \widehat{\mathbf{A}}\big]\mathbf{H}^G(t).
\label{eq: cgp euler discretization}
\end{aligned}
\end{equation}
\revision{On this basis}, we can find that Equation \ref{eq: SGC} rigidly ties the propagation depth and integration time by enforcing $\Delta t_{cgp}=1$ (i.e., the above equation degrades to Equation \ref{eq: SGC} when \revision{step size} $\Delta t_{cgp} = 1$). If so, letting $K=T_{cgp} \rightarrow \infty$ not only makes the graph Laplacian eigenvalues in a discrete propagation tend to zeros (see Appendix \ref{appx: property 1}) but also leads to infinite numerical errors (see Appendix \ref{appx: property 2}), which prevents the model from accurately capturing long-range spatial dependencies. In this work, inspired by \cite{wang2021dissecting}, we disentangle the coupling between $K$ and $T_{cgp}$, which alleviates the aforementioned problems by avoiding $T_{cgp} \rightarrow \infty$. We provide detailed theoretical justifications in Subsection \ref{subsec:theoritical comparision}. In \method, we generalize Equation \ref{eq: SGC} with its continuous formulation in the following proposition based on Equation \ref{eq: cgp euler discretization}, \revision{which allows fine-grained and long-range spatial dependencies between time series to be captured.}
\\

\textbf{Proposition 1.} \textit{The continuous dynamics of simplified graph propagation described in Equation \ref{eq: SGC} admits the following ODE:}
\begin{equation}
    \frac{\displaystyle\mathrm{d}\mathbf{H}^G(t)}{\displaystyle\mathrm{d}t}=(\widehat{\mathbf{A}}-\mathbf{I}_N) \ \mathbf{H}^G(t),
\label{eq: cgp general}
\end{equation}
\revision{\textit{where the initial state $\mathbf{H}^G(0) = \mathbf{H}^{G}_{0}$. Specifically, $\mathbf{H}^{G}_{0}$ is the intermediate state of the continuous temporal aggregation process (Subsection \ref{subsec:CTA}) as mentioned before.}} \\

To further reduce numerical errors, we propose an attentive transformation to replace the linear mapping in Equation \ref{eq: SGC}, which integrates not only the final but also the initial and selected intermediate states as the output of graph propagation:
\begin{equation}
\left\{
\begin{aligned}
&\mathbf{H}^G(t_i) = \operatorname{ODESolve}(\mathbf{H}^G(0), \frac{\displaystyle\mathrm{d}\mathbf{H}^G(t)}{\displaystyle\mathrm{d}t}, t_i),
\\
& \mathbf{H}_{out}^{G}=\sum_{t_i} \mathbf{H}^{G}(t_i) \ \mathbf{\Phi_{t_i}},\ \ t_i \in [0, T_{cgp}],
\end{aligned}
\right.
\label{eq: cgp attention}
\end{equation}
where $\operatorname{ODESolve}(\cdot)$ can be any black-box ODE solver introduced in \cite{chen2018neural}. Specifically, $\mathbf{H}^{G}(t_i)$ denotes the selected intermediate states of a CGP process, and we only take $t_i$ that is divisible by $\Delta t_{cgp}$ for simplicity in practice. \\

\noindent \textbf{Dynamic graph structure learning.} In Equation \ref{eq: cgp general}, it remains unknown how the graph adjacency matrix $\mathbf{A}$ is constructed. To address the third challenge and handle multivariate time series without graph priors (e.g., unknown $\mathbf{A}$), we adopt a direct optimization approach to learn dynamic graph structures together with the entire model, where node connections evolve with model training. Specifically, for a sequence of historical observations, the underlying adjacency matrix $\mathbf{A}$ is dynamically optimized as training progresses to learn to describe the stable interdependencies between variables:
\begin{equation}
\left\{
\begin{aligned}
    & \mathbf{M}^{k} = \tanh(\beta \ \mathbf{E}^k \mathbf{\Gamma}_{gc}^k), \ k \in \{1, 2\}, \\
    & \mathbf{A}_{ij}=\operatorname{ReLU}\Big(\tanh\big(\beta(\mathbf{M}^{1}_{ij}{\mathbf{M}^{2}_{ij}}^\mathsf{T} - \mathbf{M}^{2}_{ij}{\mathbf{M}^{1}_{ij}}^\mathsf{T})\big)\Big),
\end{aligned}
\right.
\label{eq: graph construction}
\end{equation}
where $\mathbf{M}^{1}, \mathbf{M}^{2} \in \mathbb{R}^{N \times d}$ are described by two neural networks with randomly initialized embedding matrices $\mathbf{E}^1, \mathbf{E}^2 \in \mathbb{R}^{N \times d}$ and trainable parameters $\mathbf{\Gamma}_{gc}^1, \mathbf{\Gamma}_{gc}^2 \in \mathbb{R}^{d \times d}$. $\beta$ is a hyperparameter to adjust the activation saturation rate. The learned graph structure is made sparse to reduce the computational cost and is supposed to be uni-directional because changes in a time series are likely to unidirectionally lead to fluctuations in other series \cite{wu2020connecting}.

\subsection{Continuous Temporal Aggregation}\label{subsec:CTA}
Solving the spatial ODE in Equation \ref{eq: cgp attention} only allows capturing the spatial dependencies between time series at a certain time step. To learn from the rich temporal information, we treat our spatial ODE as an interior process of the proposed temporal \revision{neural} ODE, which allows \method to model precise and stable dynamics of multivariate time series from both spatial and temporal perspectives.

We first introduce the composition of temporal \revision{neural} ODE to characterize fine-grained and accurate temporal dependencies. Given the shortcomings of RNNs, such as time-consuming iteration and gradient explosion \cite{wu2019graph}, we may stack multiple residual convolution blocks to extract and aggregate temporal patterns in a non-recursive manner:
\begin{equation}
    \mathbf{H}^{T}_{l+1} = \mathcal{T}(\mathbf{H}^{T}_{l}, Q_{l+1}) + \operatorname{TCN}(\mathbf{H}^{T}_{l}, {\mathbf{\Theta}}_{l}), \ l \in \{0, \cdots, \revision{L-1}\},
\label{eq: residual dilated tcn block}
\end{equation}
where $\operatorname{TCN}(\cdot, \mathbf{\Theta}_l)$ is an individually parameterized temporal convolution layer, $\mathcal{T}(\mathbf{H}^{T}_{l}, Q_{l+1})$ denotes a truncate function to take only the last $Q_{l+1}$ elements in $\mathbf{H}^{T}_{l}$ along its last dimension, and $\mathbf{H}^{T}_l \in \mathbb{R}^{N \times D' \times Q_l}$ is the output of the $l$-th layer with the sequence length $Q_l$. In this formulation, the last dimension of the residual input $\mathbf{H}^{T}_l$ has to be truncated to $Q_{l+1}$ before adding to its transformation because the length of latent representations shrinks gradually after each aggregation step, i.e., $Q_{l+1} = Q_{l} - r^l \times (k-1)$ and $Q_1=R-k+1$. Specifically, we define $\mathbf{H}^{T}_{0} \in \mathbb{R}^{N \times D' \times R}$ as the initial state, $r$, $k$ and $R$ are dilation factor, kernel size, and model receptive field. In practice, we assure $R>T$ to losslessly encode all historical observations, where $R=L(k-1)+1$ when $r=1$, and $R=1+(k-1)(r^L-1)/(r-1)$ when $r>1$.

However, the discrete formulation in Equation \ref{eq: residual dilated tcn block} suffers from two main limitations. Firstly, it fails to model the fine-grained and accurate temporal dynamics with a fixed large \revision{step size in numerical integration}, i.e., $\Delta t_{cta} = 1$, which breaks the continuity of the latent state trajectories. Secondly, it parameterizes convolution layers individually, which has a large number of trainable parameters and relies on dedicated model designs to avoid the gradient vanishing issue and ensure convergence \cite{wu2019graph, wu2020connecting}, resulting in high computational and memory \revision{overheads}. 
Thus, we apply a similar idea to disentangle the ties between aggregation depth $L$ and integration time $T_{cta}$ by letting $L = T_{cta} / \Delta t_{cta}$. In such a way, 
\revision{given a desired terminate time $T_{cta}$}
and initial state $\mathbf{H}^{T}_0$, we can characterize the entire continuous temporal aggregation process with a single set of parameters $\mathbf{\Theta}$ by letting $\Delta t_{cta} \rightarrow 0$:
\begin{equation}
\left\{
\begin{aligned}
    & \revision{\widetilde{\mathbf{H}}^{T}_{out}} = \mathbf{H}^{T}_{0} + \int_{\revision{0}}^{\revision{T_{cta}}} \mathcal{P}\big(\operatorname{TCN}(\mathbf{H}^{T}_{t}, \mathbf{\Theta}), R\big) \ \mathrm{d}t, \\
    & \mathbf{H}_{out}^{T} = \revision{\widetilde{\mathbf{H}}^{T}_{out}}[\cdots, -1].
\end{aligned}
\right.
    \label{eq: padding-based residual dilated tcn block v2}
\end{equation} 

To achieve this, we design a simple zero-padding trick to ensure the invariance of latent state dimensions during transformations, where the length (i.e., the last dimension) of latent states are left zero-padded to $R$ with a padding function $\mathcal{P}(\cdot)$ after each step of aggregation. \revision{Although the padding is applied, the length of informative parts of latent states shrinks gradually to one after the temporal aggregation (Figure \ref{fig:framework}), as same as in temporal convolution networks. Thus, we take $\widetilde{\mathbf{H}}^{T}_{out}[\cdots, -1]$ as the output of the proposed CTA process in Equation \ref{eq: padding-based residual dilated tcn block v2}.} On this basis, we have the second proposition defined as follows: \\

\textbf{Proposition 2} \textit{The temporal aggregation process described in Equation \ref{eq: residual dilated tcn block} is a discretization of the following ODE:}
\begin{equation}
    \frac{\displaystyle\mathrm{d}\mathbf{H}^{T}(t)}{\displaystyle\mathrm{d}t} =\mathcal{P}\big(\operatorname{TCN}(\mathbf{H}^{T}(t), t, \mathbf{\Theta}), R \big),
    \label{eq: cta general}
\end{equation}
\textit{with the initial state $\mathbf{H}^{T}(0) = \mathbf{H}^{T}_0$, which is obtained by mapping the input series to the latent space with a separate convolution layer parameterized by $\mathbf{\Gamma}_{sc}$, i.e., $\mathbf{H}^{T}_0=\operatorname{Conv_{1 \times 1}}(\mathbf{X}_{t+1:t+T}, \mathbf{\Gamma}_{sc})$. We denote this mapping as the start convolution in Figure \ref{fig:framework}.} \\

Regarding the design of $\operatorname{TCN}(\cdot, \mathbf{\Theta})$ in the above proposition, we adopt a gating mechanism to control the amount of information flows at each integration step:
\begin{equation}
    \revision{\operatorname{TCN}(\mathbf{H}^{T}(t), \mathbf{\Theta}) = f_{\mathcal{C}}(\mathbf{H}^{T}(t), \mathbf{\Theta}_c) \odot f_{\mathcal{G}}(\mathbf{H}^{T}(t), \mathbf{\Theta}_g),}
    \label{eq: dilated tcn within cta}
\end{equation}
where $\odot$ denotes the element-wise product. $f_{\mathcal{C}}(\cdot,\mathbf{\Theta}_c)$ and $f_{\mathcal{G}}(\cdot,\mathbf{\Theta}_g)$ are filtering and gating convolutions that share similar network structures but with different parameters and nonlinearities:
\begin{equation}
\left\{
\begin{aligned}
    & \revision{f_{\mathcal{C}}(\mathbf{H}^{T}(t), \mathbf{\Theta}_c)} = \tanh\big(\mathbf{W}_{\mathbf{\Theta}_{c}^{1 \times m}} \star_{\delta} \mathbf{H}^{T}(t) + \mathbf{b}_{\mathbf{\Theta}_{c}^{1 \times m}} \big), \\
    & \revision{f_{\mathcal{G}}(\mathbf{H}^{T}(t), \mathbf{\Theta}_g)} = \sigma\big(\mathbf{W}_{\mathbf{\Theta}_{g}^{1 \times m}} \star_{\delta} \mathbf{H}^{T}(t) + \mathbf{b}_{\mathbf{\Theta}_{g}^{1 \times m}} \big),
\end{aligned}
\right.
    \label{eq: filter network}
\end{equation}
where $\sigma(\cdot)$ represents the sigmoid activation, and $\star_{\delta}$ denotes the convolution operation with an expandable dilation defined by $\delta = \floor{r^{t/\Delta t_{cta}}}$. \revision{Specifically, $\lfloor \cdot {\rfloor}$ denotes the floor operation, which outputs the largest integer less than or equal to the input.}
In practice, adopting a single kernel size is less effective in exploring multi-granularity temporal patterns. Thus, inspired by \cite{wu2020connecting}, we equip $f_{\mathcal{C}}(\cdot,\mathbf{\Theta}_c)$ and $f_{\mathcal{G}}(\cdot,\mathbf{\Theta}_g)$ with multiple convolutions with different kernel widths $m$, i.e., $f_{\mathcal{C}}(\cdot, \mathbf{\Theta}_{c}^{1 \times m})$ and $f_{\mathcal{G}}(\cdot, \mathbf{\Theta}_{g}^{1 \times m})$. Since most of the time series data have inherent periods (e.g., 7, 14, 24, 28, and 30), letting kernel width in set $\{2,3,6,7\}$ makes the aforementioned periods can be fully covered.

\begin{algorithm}[t]
	\caption{The Training Algorithm of \method}
	\label{algo: overall algorithm}
	\vspace{1mm}
    \textbf{Input}: Training data $\mathbf{X}$, input length $T$, forecasting horizon $H$, batch size $B$, training epoch $E$, learning rate $\eta$, and the initialized \method model $F(\cdot)$ with $\mathbf{\Theta}$, $\mathbf{\Phi}$, and $\mathbf{\Gamma}$. \\
    \textbf{Output}: Well-trained \method model $F^*(\cdot)$. \\ \vspace{-4mm}
    \begin{algorithmic}[1]
		\STATE $\text{data} \leftarrow \operatorname{DataLoader}(\mathbf{X}, T, H, B)$;
		\FOR{$i \in 1,2,\cdots,E$}
			\STATE $\text{data}.\operatorname{shuffle}()$;
			\FOR{$(\mathbf{\mathcal{X}}, \mathbf{\mathcal{Y}})$ in $\operatorname{enumerate}\big(\text{data}.\operatorname{get\_iterator}()\big)$}
			\STATE $/*$ {\it See Algorithm \ref{algo: solving MTGODE} for details} $*/$
			\STATE $\widehat{\mathbf{\mathcal{Y}}} \leftarrow F(\mathbf{\mathcal{X}}; \mathbf{\Theta}, \mathbf{\Phi}, \mathbf{\Gamma})$;
			\STATE $\ell \leftarrow \text{MAE}(\widehat{\mathbf{\mathcal{Y}}}, \mathbf{\mathcal{Y}})$;
			\STATE Calculate the stochastic gradients of $\mathbf{\Theta}$, $\mathbf{\Phi}$, and $\mathbf{\Gamma}$ w.r.t. $\ell$;
            \STATE Update $\mathbf{\Theta}$, $\mathbf{\Phi}$, and $\mathbf{\Gamma}$ w.r.t. their gradients and $\eta$;
            \ENDFOR
			\STATE $\eta \leftarrow \operatorname{LRScheduler}(\eta, i)$;
		\ENDFOR
	\end{algorithmic}
\end{algorithm}

\begin{algorithm}[t]
	\caption{Solving the \method $F(\cdot)$}
	\label{algo: solving MTGODE}
    \textbf{Input}: Input multivariate time series $\mathbf{\mathcal{X}}$, temporal terminal time $T_{cta}$, spatial terminal time $T_{cgp}$, and dilation factor $r$. \\
    \textbf{Output}: The forecasting results $\widehat{\mathbf{\mathcal{Y}}}$. \\ \vspace{-4mm}
    \begin{algorithmic}[1]
          \vspace{2mm}
          \PROCEDURE{spatial\_func}{$\mathbf{H}_{in}, \mathbf{A}$}
          \STATE $/*$ {\it The spatial ODE function defined in Eq. \ref{eq: cgp general}} $*/$
          \STATE $\widehat{\mathbf{A}} \leftarrow \operatorname{adj\_norm}(\mathbf{A})$
          \STATE $\mathbf{H}_{out} \leftarrow \widehat{\mathbf{A}}\mathbf{H}_{in} - \mathbf{H}_{in}$
          \STATE \textbf{return} \ $\mathbf{H}_{out}$
          \ENDPROCEDURE 
          \vspace{2mm}
          \PROCEDURE{temporal\_func}{$\mathbf{H}_{in}$}
          \STATE $/*$ {\it The temporal convolution defined in Eq. \ref{eq: dilated tcn within cta}} $*/$
          \STATE $\widetilde{\mathbf{H}} \leftarrow \operatorname{TCN}(\mathbf{H}_{in}, \mathbf{\Theta})$
          \STATE $/*$ {\it Solve the interior ODE} $*/$
          \STATE $\mathbf{H}_0, \cdots, \mathbf{H}_{T_{cgp}} \leftarrow \operatorname{ODESolve}(\widetilde{\mathbf{H}}, spatial\_func, $\par$
          \hspace{42.5mm}\hskip\algorithmicindent 0, \cdots, T_{cgp})$
          \STATE $/*$ {\it Attentive transformation defined in Eq. \ref{eq: cgp attention}} $*/$
          \STATE $\mathbf{H} \leftarrow \operatorname{Attn}(\mathbf{H}_{0}, \cdots, \mathbf{H}_{T_{cgp}}; \mathbf{\Phi})$
          \STATE $/*$ {\it The padding trick in Eq. \ref{eq: dyg-ode function}} $*/$
          \STATE $\mathbf{H}_{out} \leftarrow \operatorname{zero\_padding}(\mathbf{H}, R)$
          \STATE $\operatorname{TCN(\cdot, \mathbf{\Theta}).update\_dilation}(r)$
          \STATE \textbf{return} \ $\mathbf{H}_{out}$
          \ENDPROCEDURE
          \vspace{2mm}
          \STATE $/*$ {\it Map input to the latent space} $*/$
          \STATE $\mathbf{H}_0 \leftarrow \operatorname{Conv}(\mathcal{X}, \mathbf{\Gamma_{sc}})$ 
          \STATE $/*$ {\it Dynamically acquire the underlying graph structure} $*/$
          \STATE $node\_idx \leftarrow \operatorname{list}\big(\operatorname{range}(\mathcal{X}.shape[1])\big)$ 
          \STATE $\mathbf{A} \leftarrow \operatorname{graph\_learner}\big(node\_idx, \mathbf{\Gamma_{gc}}\big)$ 
          \STATE $/*$ {\it Solve the exterior ODE defined in Eq. \ref{eq: dyg-ode function}} $*/$
          \STATE $\mathbf{H} \leftarrow \operatorname{ODESolver}(\mathbf{H}_0, \revision{temporal\_func}, T_{cta})$ 
          \STATE $/*$ {\it Downstream decoding} $*/$
          \STATE $\widehat{\mathcal{Y}} \leftarrow \operatorname{Conv}(\mathbf{H}, \mathbf{\Gamma_{dc}})$ 
	\end{algorithmic}
\end{algorithm}

\subsection{Comparison with Discrete Variants}\label{subsec:theoritical comparision}
Compared with the existing GNN-based methods \cite{yu2018spatio, wu2019graph, wu2020connecting}, our approach is free from the over-smoothing issue (e.g., the model performance drops when the depth of graph propagation increases). This allows the CGP process to capture stable long-range spatial dependencies by disengaging the ties between graph propagation depth and integration time. 
Specifically, our method possesses the following properties: \\

\textbf{Property 1} \textit{\revision{For a specific integration time $T_{cgp}$, \method ensures the convergence of learned spatial representations by letting $K=T_{cgp}/ \Delta t_{cgp}$ and $K \rightarrow \infty$.}} \\

\textit{Proof.} See Appendix \ref{appx: property 1}. \qed \\

\textbf{Property 2} \textit{\revision{In \method, letting $K=T_{cgp}/ \Delta t_{cgp} \rightarrow \infty$ makes the numerical errors of spatial modeling approaching zero with a fixed integration time $T_{cgp}$.}} \\

\textit{Proof.} See Appendix \ref{appx: property 2}. \qed \\

The above properties are further empirically validated in Subsection \ref{subsec:study on two ODEs}. Similarly, compared with the discrete temporal convolutions in existing works, the proposed temporal \revision{neural} ODE directly parameterizes the derivation of latent states to characterize the nature of temporal information aggregation. In such a way, the ties between aggregation depth, integration time, and memory bottleneck are disentangled, as we explained in Subsection \ref{subsec:CTA}. This allows the CTA process to effectively and efficiently learn fine-grained and more accurate temporal dynamics with a single set of parameters by shrinking the \revision{step size} $\Delta t_{cta}$ under a specific \revision{integral interval $[0, T_{cta}]$.}

\subsection{Overall Architecture and Model Training}\label{subsec:model training}
\noindent \textbf{Overall architecture.} 
We have the proposed \method method defined below by unifying the proposed continuous spatial and temporal message-passing mechanisms.
It is worth noting that instead of simply concatenating them end-to-end, we take each intermediate state of the \textit{exterior} CTA process as the initial state of \textit{interior} CGP process,
thereby allowing the model to characterize the underlying interlaced spatial-temporal dynamics of input series in a fully continuous manner to derive more expressive representations for downstream forecasting tasks.
Given two black-box ODE solvers, i.e., $\operatorname{ODESolve}^1(\cdot)$ and $\operatorname{ODESolve}^2(\cdot)$, the learned spatial-temporal representations of input series can be obtained by integrating $\displaystyle\mathrm{d}\mathbf{H}(t) / \displaystyle\mathrm{d}t$:
\begin{equation}
    \mathbf{H}_{out} = \operatorname{ODESolve}^1\big(\mathbf{H}(0), \frac{\displaystyle\mathrm{d}\mathbf{H}(t)}{\displaystyle\mathrm{d}t}, T_{cta} \big),
\label{eq: dyg-ode}
\end{equation}
where we have $\displaystyle\mathrm{d}\mathbf{H}(t) / \displaystyle\mathrm{d}t$ defined below based on the aforementioned two propositions:
\begin{equation}
    \begin{aligned}
    \frac{\displaystyle\mathrm{d}\mathbf{H}(t)}{\displaystyle\mathrm{d}t} =\mathcal{P}\bigg( \mathcal{A}\Big(&\operatorname{ODESolve}^2\big( \operatorname{TCN}(\mathbf{H}(t), t, \mathbf{\Theta}), \\
    &\ \ \frac{\displaystyle\mathrm{d}\mathbf{H}^{G}(\tau)}{\displaystyle\mathrm{d}\tau}, 0, \cdots, T_{cgp} \big), \mathbf{\Phi} \Big), R \bigg).
    \end{aligned}
\label{eq: dyg-ode function}
\end{equation}

In the above equations, the interior ODE solving and the attentive transformation, i.e., $\mathcal{A}(\cdot, \mathbf{\Phi})$, are given by Equation \ref{eq: cgp attention} by letting $\mathbf{H}^G(0)=\operatorname{TCN}(\mathbf{H}(t), t, \mathbf{\Theta})$. 
In particular, we let the initial state $\mathbf{H}(0) = \mathbf{H}^{T}(0)$ in Equation \ref{eq: cta general}, and further define two black-box ODE solvers as the Euler or Runge-Kutta method adopted in \cite{chen2018neural} with different selected integration time and \revision{step sizes} for simplicity. \\

\noindent \textbf{Model training.} Given a sequence of historical observations $\mathbf{X}_{t+1:t+T}$, we first learn its representation $\mathbf{H}_{out} \in \mathbb{R}^{N \times D'}$ via Equation \ref{eq: dyg-ode} and then make the forecasting with a downstream decoder $g(\cdot, \mathbf{\Gamma}_{dc})$, as the end convolution shown in Figure \ref{fig:framework}. Thus, our training objective described in Equation \ref{eq: problem} can be reformulated as follows:
\begin{equation}
 f^*, g^* = \mathop{\arg\min}\limits_{f, g}\sum_{t}\ell\Big(g\big(f(\mathbf{X}_{t+1:t+T}; \mathbf{\Theta}, \mathbf{\Phi}, \revision{\mathbf{\Gamma^{'}}}),
\mathbf{\Gamma}_{dc} \big), \mathbf{Y}\Big),
\label{eq: final objective}
\end{equation}
where \revision{$\mathbf{\Gamma^{'}}=\{\mathbf{\Gamma}_{sc},\mathbf{\Gamma}_{gc} \}$}, and $\ell(\cdot)$ denotes the mean absolute error (MAE).
We illustrate the optimization of \method in Algorithm \ref{algo: overall algorithm} and \ref{algo: solving MTGODE}. \\

\noindent \textbf{Complexity Analysis.}
We analyze the time complexity of the proposed method. For the dynamic graph structure learning module, the time complexity is $\mathcal{O}(Nd^2+N^2d)$, where $N$ and $d$ are the number of variables and the dimension of embedding matrices in Equation \ref{eq: graph construction}. For a single-step graph propagation, the time complexity is $\mathcal{O}(ED'+ND'^2)$, where $E$ and $D'$ are the number of edges and latent state dimensions. 
Thus, the time complexity of the proposed CGP module is $\mathcal{O}\big(T_{cgp}/\Delta t_{cgp}(ED'+ND'^2)\big)$. 
For a single-step temporal aggregation, the time complexity is $\mathcal{O}(NQ_lD'^2/r)$. We use $Q_l$ and $r$ to denote the length of latent states at $l$-th layer in the temporal module and the dilation factor of $\text{TCN}(\cdot, \mathbf{\Theta})$. According to the padding trick mentioned in Equation \ref{eq: padding-based residual dilated tcn block v2}, the time complexity of the proposed CTA module is $\mathcal{O}\big(T_{cta}/\Delta t_{cta}(NRD'^2/r)\big)$, where $R$ denotes the temporal reception field of \method discussed in Subsection \ref{subsec:CTA}. 
Compared with the discrete counterparts of our method, e.g., MTGNN \cite{wu2020connecting}, our model is less complex by eliminating redundant neural connections, such as the residual and skip layers with the time complexity of $\mathcal{O}(LNRD'^2)$, where $L$ denotes the number of layers in discrete models.
In the following section, we further evaluate the model efficiency of \method empirically from various perspectives \revision{to demonstrate its superiority.}

\section{Experimental Study}\label{sec:experiment}
In this section, we conduct comprehensive experiments on five real-world benchmark datasets to show the performance of \method.
We compare our method with the state-of-the-art time series forecasting methods and follow their configurations to conduct experiments for fair comparisons. 
In this section, we also empirically demonstrate the potency and efficiency of two proposed continuous regimes, showing superior properties compared with discrete variants.
Also, ablation and parameter sensitivity studies are conducted to further investigate the properties of \method.

\begin{table}[ht]
	\centering
	\caption{The statistics of five multivariate time series datasets.}
	\begin{tabular}{@{}c|c|c|c|c@{}}
		\toprule
		\textbf{Dataset} & \textbf{\# Samples} & \textbf{\# Nodes} & \textbf{\makecell[c]{Sampling \\ Rate}} & \textbf{\makecell[c]{Predefined \\ Graph}} \\
		\midrule
		\textbf{Electricity} & 26,304 & 321 & 1 hrs & No \\
		\textbf{Solar-Energy} & 52,560  & 137  & 10 mins & No \\
		\textbf{Traffic} & 17,544 & 862  & 1 hrs  & No \\ 
		\midrule
		\textbf{Metr-La} & 34,272 & 207 & 5 mins & Yes \\
		\textbf{Pems-Bay} & 52,116 & 4,732 & 5 mins & Yes \\
		\bottomrule
	\end{tabular}
	\label{table:dataset}
\end{table}

\subsection{Dataset Description}\label{subsec:datasets}
We experiment on five benchmark datasets to evaluate the performance of \method and its competitors. Three of these are conventional time series datasets \cite{lai2018modeling}, i.e., Electricity, Solar-Energy, and Traffic, without predefined graph structures, and the rest two are traffic datasets \cite{li2018diffusion}, i.e., Metr-La and Pems-Bay, with predefined sensor maps (i.e., graph structures).
We summarize the dataset statistics in Table \ref{table:dataset} and provide a detailed description of them as follows:
\begin{itemize}
    \item \textbf{Electricity\footnote{\url{https://github.com/laiguokun/multivariate-time-series-data} \label{fn: electricity dataset}}}: This dataset consists of the energy consumption records of 321 clients between 2012 and 2014 with the sampling rate set to 1 hour.
    \item \textbf{Solar-Energy\textsuperscript{\ref{fn: electricity dataset}}}: It contains the solar power production records of 137 PV plants in Alabama State in the year 2006, where the sampling rate is 10 minutes.
    \item \textbf{Traffic\textsuperscript{\ref{fn: electricity dataset}}}: A collection of hourly road occupancy rates measured by 862 sensors in the San Francisco Bay area between 2015 and 2016.
    \item \textbf{Metr-La\footnote{\url{https://github.com/liyaguang/DCRNN} \label{fn: metr-la dataset}}}: It contains the traffic speed readings with 5 minutes sampling rate from the 207 loop detectors in Los Angeles County highways in the year of 2012.
    \item \textbf{Pems-Bay\textsuperscript{\ref{fn: metr-la dataset}}}: This dataset is provided by California Transportation Agencies Performance Measurement Systems, which consists of the traffic speed readings of 325 sensors in the Bay Area in the year 2017, where the data sampling rate is same as in Metr-La.
\end{itemize}

\begin{table*}[htbp]
    \small
    \centering
 	\caption{Single-step forecasting results on three benchmark time series datasets. \colorbox[HTML]{ececec}{\textbf{Bold}} denotes the best performances. \revision{OOM indicates out-of-memory.}}
    \begin{NiceTabular}{c| c c| c | p{35 pt}<{\centering} p{35 pt}<{\centering} p{35 pt}<{\centering} p{35 pt}<{\centering} p{35 pt}<{\centering} p{35 pt}<{\centering} p{35 pt}<{\centering} p{35 pt}<{\centering}}
		
		\toprule
		
		\Block{1-3}{Dataset} &  &  & \ \ Metric \ \ & VARMLP & GRU & LSTNet & TPA-LSTM & MTGNN & HyDCNN & \revision{STG-NCDE} & MTGODE \textbf{(Ours)} \\
		\bottomrule

		\multirow{6}{*}
		{\rotatebox{90}{Electricity}} & \multirow{6}{*}
		{\rotatebox{90}{Horizon}} & \multirow{2}{*}
		{3} & RSE$\downarrow$ & 0.1393 & 0.1102 & 0.0864 & 0.0823 & 0.0745 & 0.0832 & \revision{0.6152} & \cellcolor[HTML]{ececec}\textcolor{Red}{\textbf{0.0736}} \\
		
		 &  &  & CORR$\uparrow$ & 0.8708 & 0.8597 & 0.9283 & 0.9439 & \cellcolor[HTML]{ececec}\textcolor{Red}{\textbf{0.9474}} & 0.9354 & \revision{0.8739} & 0.9430 \\ \cmidrule{4-12}
		
		 &  & \multirow{2}{*}
		{6} & RSE$\downarrow$ & 0.1620 & 0.1144 & 0.0931 & 0.0916 & 0.0878 & 0.0898 & \revision{0.6584} & \cellcolor[HTML]{ececec}\textcolor{Red}{\textbf{0.0809}} \\
		
		\multicolumn{1}{l}{} & \multicolumn{1}{l}{} & \multicolumn{1}{l}{} & CORR$\uparrow$ & 0.8389 & 0.8623 & 0.9135 & 0.9337 & 0.9316 & 0.9329 & \revision{0.8663} & \cellcolor[HTML]{ececec}\textcolor{Red}{\textbf{0.9340}} \\ \cmidrule{4-12}
		
		\multicolumn{1}{l}{} & \multicolumn{1}{l}{} & \multicolumn{1}{c}{\multirow{2}{*}
		{12}} & RSE$\downarrow$ & 0.1557 & 0.1183 & 0.1007 & 0.0964 & 0.0916 & 0.0921 & \revision{0.7302} & \cellcolor[HTML]{ececec}\textcolor{Red}{\textbf{0.0891}} \\
		
		\multicolumn{1}{l}{} & \multicolumn{1}{l}{} & \multicolumn{1}{l}{} & CORR$\uparrow$ & 0.8192 & 0.8472 & 0.9077 & 0.9250 & 0.9278 & \cellcolor[HTML]{ececec}\textcolor{Red}{\textbf{0.9285}} & \revision{0.8728} & 0.9279 \\ \midrule \midrule
		
		\multicolumn{1}{c}{\multirow{6}{*}
		{\rotatebox{90}{Traffic}}} & \multicolumn{1}{c}{\multirow{6}{*}
		{\rotatebox{90}{Horizon}}} & \multicolumn{1}{c}{\multirow{2}{*}
		{3}} & RSE$\downarrow$ & 0.5582 & 0.5358 & 0.4777 & 0.4487 & 0.4162 & 0.4198 & \revision{OOM} & \cellcolor[HTML]{ececec}\textcolor{Red}{\textbf{0.4127}} \\
		
		\multicolumn{1}{l}{} & \multicolumn{1}{l}{} & \multicolumn{1}{l}{} & CORR$\uparrow$ & 0.8245 & 0.8511 & 0.8721 & 0.8812 & 0.8963 & 0.8915 & \revision{OOM} & \cellcolor[HTML]{ececec}\textcolor{Red}{\textbf{0.9020}} \\ \cmidrule{4-12}
		
		\multicolumn{1}{l}{} & \multicolumn{1}{l}{} & \multicolumn{1}{c}{\multirow{2}{*}
		{6}} & RSE$\downarrow$ & 0.6579 & 0.5522 & 0.4893 & 0.4658 & 0.4754 & 0.4290 & \revision{OOM} & \cellcolor[HTML]{ececec}\textcolor{Red}{\textbf{0.4259}} \\
		
		\multicolumn{1}{l}{} & \multicolumn{1}{l}{} & \multicolumn{1}{l}{} & CORR$\uparrow$ & 0.7695 & 0.8405 & 0.8690 & 0.8717 & 0.8667 & 0.8855 & \revision{OOM} & \cellcolor[HTML]{ececec}\textcolor{Red}{\textbf{0.8945}} \\ \cmidrule{4-12}
		
		\multicolumn{1}{l}{} & \multicolumn{1}{l}{} & \multicolumn{1}{c}{\multirow{2}{*}
		{12}} & RSE$\downarrow$ & 0.6023 & 0.5562 & 0.4950 & 0.4641 & 0.4461 & 0.4352 & \revision{OOM} & \cellcolor[HTML]{ececec}\textcolor{Red}{\textbf{0.4329}} \\
		
		\multicolumn{1}{l}{} & \multicolumn{1}{l}{} & \multicolumn{1}{l}{} & CORR$\uparrow$ & 0.7929 & 0.8345 & 0.8614 & 0.8717 & 0.8794 & 0.8858 & \revision{OOM} & \cellcolor[HTML]{ececec}\textcolor{Red}{\textbf{0.8899}} \\ \midrule \midrule
		
		\multicolumn{1}{c}{\multirow{6}{*}
		{\rotatebox{90}{Solar-Energy}}} & \multicolumn{1}{c}{\multirow{6}{*}
		{\rotatebox{90}{Horizon}}} & \multicolumn{1}{c}{\multirow{2}{*}
		{3}} & RSE$\downarrow$ & 0.1922 & 0.1932 & 0.1843 & 0.1803 & 0.1778 & 0.1806 & \revision{0.2346} & \cellcolor[HTML]{ececec}\textcolor{Red}{\textbf{0.1693}} \\
		
		\multicolumn{1}{l}{} & \multicolumn{1}{l}{} & \multicolumn{1}{l}{} & CORR$\uparrow$ & 0.9829 & 0.9823 & 0.9843 & 0.9850 & 0.9852 & 0.9865 & \revision{0.9748} & \cellcolor[HTML]{ececec}\textcolor{Red}{\textbf{0.9868}} \\ \cmidrule{4-12}
		
		\multicolumn{1}{l}{} & \multicolumn{1}{l}{} & \multicolumn{1}{c}{\multirow{2}{*}
		{6}} & RSE$\downarrow$ & 0.2679 & 0.2628 & 0.2559 & 0.2347 & 0.2348 & 0.2335 & \revision{0.2908} & \cellcolor[HTML]{ececec}\textcolor{Red}{\textbf{0.2171}} \\
		
		\multicolumn{1}{l}{} & \multicolumn{1}{l}{} & \multicolumn{1}{l}{} & CORR$\uparrow$ & 0.9655 & 0.9675 & 0.9690 & 0.9742 & 0.9726 & 0.9747 & \revision{0.9605} & \cellcolor[HTML]{ececec}\textcolor{Red}{\textbf{0.9771}} \\ \cmidrule{4-12}
		
		\multicolumn{1}{l}{} & \multicolumn{1}{l}{} & \multicolumn{1}{c}{\multirow{2}{*}
		{12}} & RSE$\downarrow$ & 0.4244 & 0.4163 & 0.3254 & 0.3234 & 0.3109 & 0.3094 & \revision{0.5149} & \cellcolor[HTML]{ececec}\textcolor{Red}{\textbf{0.2901}} \\
		
		\multicolumn{1}{l}{} & \multicolumn{1}{l}{} & \multicolumn{1}{l}{} & CORR$\uparrow$ & 0.9058 & 0.9150 & 0.9467 & 0.9487 & 0.9509 & 0.9515 & \revision{0.8639} & \cellcolor[HTML]{ececec}\textcolor{Red}{\textbf{0.9577}} \\
		
		\bottomrule
		
	\end{NiceTabular}
	\label{table: single-step forecasting results}
\end{table*}

\subsection{Experimental Setup}\label{subsec:exp_setup}
In this subsection, we illustrate the detailed experimental setups, including baseline methods, evaluation protocols, and hyperparameter settings for replications. \\

\noindent \textbf{Baselines.} We evaluate and compare \method with representative and state-of-the-art time series baselines, such as LSTNet \cite{lai2018modeling} and HyDCNN \cite{li2021modeling}, on three time series datasets for \textit{single-step forecasting}. We further compare with strong GNN-based forecasting baselines on two traffic datasets for \textit{multi-step forecasting}, e.g., MRA-BCGN \cite{chen2020multi}, GMAN \cite{zheng2020gman}, and MTGNN \cite{wu2020connecting}. Note that our method and MTGNN do not rely on predefined graph structures so they are applicable and compared in both single/multi-step forecasting settings. We briefly introduce primary baselines as follows:
\begin{itemize} 
    
    
    \item \textbf{LSTNet}\cite{lai2018modeling}: It combines convolution and recurrent neural networks to capture the short-term and long-term multivariate temporal dependencies.
    
    \item \textbf{TPA-LSTM}\cite{shih2019temporal}: An attention-based recurrent neural network for multivariate time series forecasting.
    
    \item \textbf{HyDCNN}\cite{li2021modeling}: It forecasts time series with position-aware dilated temporal convolutions.
    
    \item \textbf{DCRNN}\cite{lai2018modeling}: A graph diffusion-based gated recurrent neural network~\cite{cho2014learning} for traffic forecasting.
    
    \item \textbf{STGCN}\cite{yu2018spatio}: It stacks graph and temporal convolutions to capture spatial and temporal patterns jointly.
    
    \item \textbf{Graph WaveNet}\cite{wu2019graph}: It is similar to STGCN but consists of graph and dilated temporal convolutions.
    
    \begin{figure}[htbp]
    \centering
    \subfigure{
    \includegraphics[width=.95\linewidth]{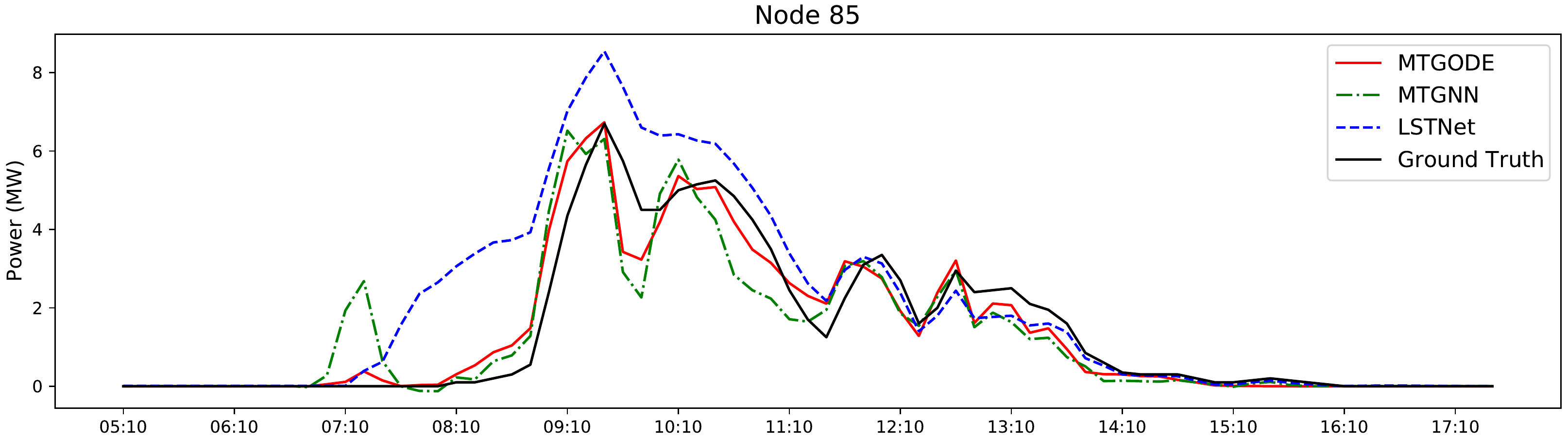}
    \label{subfig: solar-sensor15}
    }
    \subfigure{
    \includegraphics[width=.95\linewidth]{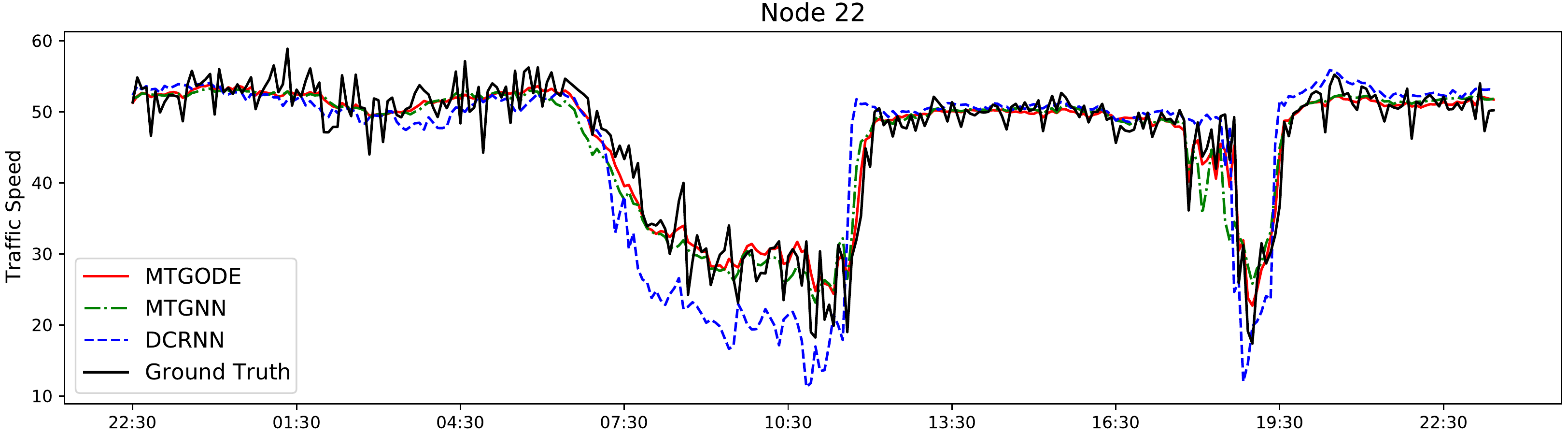}
    \label{subfig: metrla-node22}
    }
    \caption{
    Forecasting visualizations of two specific time series on Solar-Energy and Metr-La datsets.
    }
    \label{fig: visualization}
    \end{figure}
    
    \item \textbf{GMAN}\cite{zheng2020gman}: A spatial-temporal graph neural network equipped with spatial and temporal attention.

    \item \textbf{MRA-BCGN}\cite{chen2020multi}: A multi-range attentive bicomponent graph neural network for traffic forecasting. 

    \item \textbf{MTGNN}\cite{wu2020connecting}: A forecasting model based on graph neural networks and dilated temporal convolutions.
    
    \item \textbf{STGODE}\cite{fang2021spatial}: An ODE-based spatial-temporal graph neural network for traffic forecasting.
    
    \item \revision{\textbf{STG-NCDE}\cite{choi2022STGNCDE}: A spatial-temporal graph neural network based on NCDEs for traffic forecasting.}
    
\end{itemize}

\begin{table*}[t]
    \small
    \centering
 	\caption{Multi-step forecasting results on two benchmark traffic datasets. \colorbox[HTML]{ececec}{\textbf{Bold}} denotes the best results.}
	\begin{NiceTabular}{c | c | c | p{32 pt}<{\centering} p{32 pt}<{\centering} p{32 pt}<{\centering} p{32 pt}<{\centering} p{32 pt}<{\centering} p{32 pt}<{\centering} p{32 pt}<{\centering} p{32 pt}<{\centering} p{34 pt}<{\centering}}
		\toprule
		
		\Block{1-2}{Dataset} &  & Metric & DCRNN & STGCN & Graph WaveNet & GMAN & MRA-BCGN & MTGNN & STGODE & \revision{STG-NCDE} & MTGODE \textbf{(Ours)} \\
		
		\bottomrule

		\multicolumn{1}{c}{\multirow{9}{*}
		{\rotatebox{90}{Metr-La}}} & \multicolumn{1}{l}{\multirow{3}{*}
		{\rotatebox{90}{15 mins}}} & MAE & 2.77 & 2.88 & 2.69 & 2.77 & 2.67 & 2.69 & 3.47 & \revision{3.77} & \cellcolor[HTML]{ececec}\textcolor{Red}{\textbf{2.66}} \\
		
		\multicolumn{1}{l}{} & \multicolumn{1}{l}{} & RMSE & 5.38 & 5.74 & 5.15 & 5.48 & 5.12 & 5.18 & 6.76 & \revision{9.47} & \cellcolor[HTML]{ececec}\textcolor{Red}{\textbf{5.10}} \\
		
		\multicolumn{1}{l}{} & \multicolumn{1}{l}{} & MAPE (\%) & 7.30 & 7.62 & 6.90 & 7.25 & \cellcolor[HTML]{ececec}\textcolor{Red}{\textbf{6.80}} & 6.90 & 8.76 & \revision{8.54} & 6.87 \\ \cmidrule{2-12}
		
		\multicolumn{1}{l}{} & \multicolumn{1}{l}{\multirow{3}{*}
		{\rotatebox{90}{30 mins}}} & MAE & 3.15 & 3.47 & 3.07 & 3.07 & 3.06 & 3.05 & 4.36 & \revision{4.84} & \cellcolor[HTML]{ececec}\textcolor{Red}{\textbf{3.00}} \\
		
		\multicolumn{1}{l}{} & \multicolumn{1}{l}{} & RMSE & 6.45 & 7.24 & 6.22 & 6.34 & 6.17 & 6.18 & 8.47 & \revision{12.04} & \cellcolor[HTML]{ececec}\textcolor{Red}{\textbf{6.05}} \\
		
		\multicolumn{1}{l}{} & \multicolumn{1}{l}{} & MAPE (\%) & 8.80 & 9.57 & 8.37 & 8.35 & 8.30 & 8.21 & 11.14 & \revision{10.63} & \cellcolor[HTML]{ececec}\textcolor{Red}{\textbf{8.19}} \\ \cmidrule{2-12}
		
		\multicolumn{1}{l}{} & \multicolumn{1}{l}{\multirow{3}{*}
		{\rotatebox{90}{60 mins}}} & MAE & 3.60 & 4.59 & 3.53 & 3.40 & 3.49 & 3.50 & 5.50 & \revision{6.35} & \cellcolor[HTML]{ececec}\textcolor{Red}{\textbf{3.39}} \\
		
		\multicolumn{1}{l}{} & \multicolumn{1}{l}{} & RMSE & 7.60 & 9.40 & 7.37 & 7.21 & 7.30 & 7.25 & 10.33 & \revision{14.94} & \cellcolor[HTML]{ececec}\textcolor{Red}{\textbf{7.05}} \\
		
		\multicolumn{1}{l}{} & \multicolumn{1}{l}{} & MAPE (\%) & 10.5 & 12.7 & 10.01 & \cellcolor[HTML]{ececec}\textcolor{Red}{\textbf{9.72}} & 10.00 & 9.90 & 14.32 & \revision{13.49} & 9.80 \\ \midrule \midrule
		
		\multicolumn{1}{c}{\multirow{9}{*}
		{\rotatebox{90}{Pems-Bay}}} & \multicolumn{1}{l}{\multirow{3}{*}
		{\rotatebox{90}{15 mins}}} & MAE & 1.38 & 1.36 & 1.30 & 1.34 & 1.29 & 1.34 & 1.43 & \revision{1.38} & \cellcolor[HTML]{ececec}\textcolor{Red}{\textbf{1.29}} \\
		
		\multicolumn{1}{l}{} & \multicolumn{1}{l}{} & RMSE & 2.95 & 2.96 & 2.74 & 2.82 & \cellcolor[HTML]{ececec}\textcolor{Red}{\textbf{2.72}} & 2.81 & 2.88 & \revision{2.93} & 2.73 \\
		
		\multicolumn{1}{l}{} & \multicolumn{1}{l}{} & MAPE (\%) & 2.90 & 2.90 & 2.73 & 2.81 & 2.90 & 2.82 & 2.99 & \revision{2.91} & \cellcolor[HTML]{ececec}\textcolor{Red}{\textbf{2.72}} \\ \cmidrule{2-12}
		
		\multicolumn{1}{l}{} & \multicolumn{1}{l}{\multirow{3}{*}
		{\rotatebox{90}{30 mins}}} & MAE & 1.74 & 1.81 & 1.63 & 1.62 & 1.61 & 1.66 & 1.84 & \revision{1.71} & \cellcolor[HTML]{ececec}\textcolor{Red}{\textbf{1.61}} \\
		
		\multicolumn{1}{l}{} & \multicolumn{1}{l}{} & RMSE & 3.97 & 4.27 & 3.70 & 3.72 & 3.67 & 3.74 & 3.90 & \revision{3.84} & \cellcolor[HTML]{ececec}\textcolor{Red}{\textbf{3.66}} \\
		
		\multicolumn{1}{l}{} & \multicolumn{1}{l}{} & MAPE (\%) & 3.90 & 4.17 & 3.67 & 3.63 & 3.80 & 3.72 & 3.84 & \revision{3.91} & \cellcolor[HTML]{ececec}\textcolor{Red}{\textbf{3.61}} \\ \cmidrule{2-12}
		
		\multicolumn{1}{l}{} & \multicolumn{1}{l}{\multirow{3}{*}
		{\rotatebox{90}{60 mins}}} & MAE & 2.07 & 2.49 & 1.95 & \cellcolor[HTML]{ececec}\textcolor{Red}{\textbf{1.86}} & 1.91 & 1.94 & 2.30 & \revision{2.03} & 1.88 \\
		
		\multicolumn{1}{l}{} & \multicolumn{1}{l}{} & RMSE & 4.74 & 5.69 & 4.52 & 4.32 & 4.46 & 4.48 & 4.89 & \revision{4.58} & \cellcolor[HTML]{ececec}\textcolor{Red}{\textbf{4.31}} \\
		
		\multicolumn{1}{l}{} & \multicolumn{1}{l}{} & MAPE (\%) & 4.90 & 5.79 & 4.63 & \cellcolor[HTML]{ececec}\textcolor{Red}{\textbf{4.31}} & 4.60 & 4.58 & 4.61 & \revision{4.82} & 4.39 \\

		\bottomrule
		
	\end{NiceTabular}
    \label{table: multi-step forecasting results}
\end{table*}

\begin{figure*}[htbp]
\centering
\subfigure{
\includegraphics[width=5.5cm]{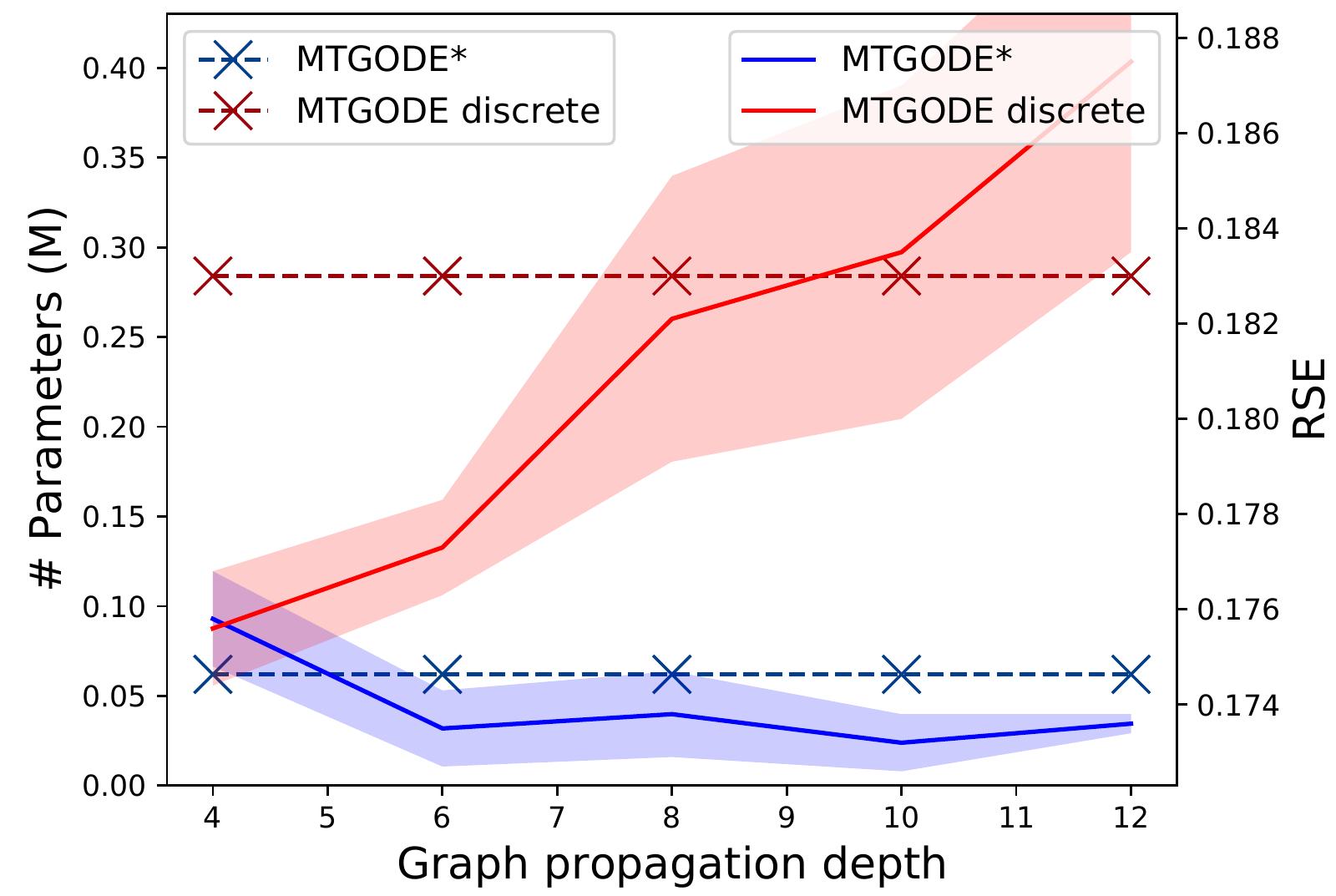} 
\label{subfig: graph propagation depth}
}
\subfigure{
\includegraphics[width=5.5cm]{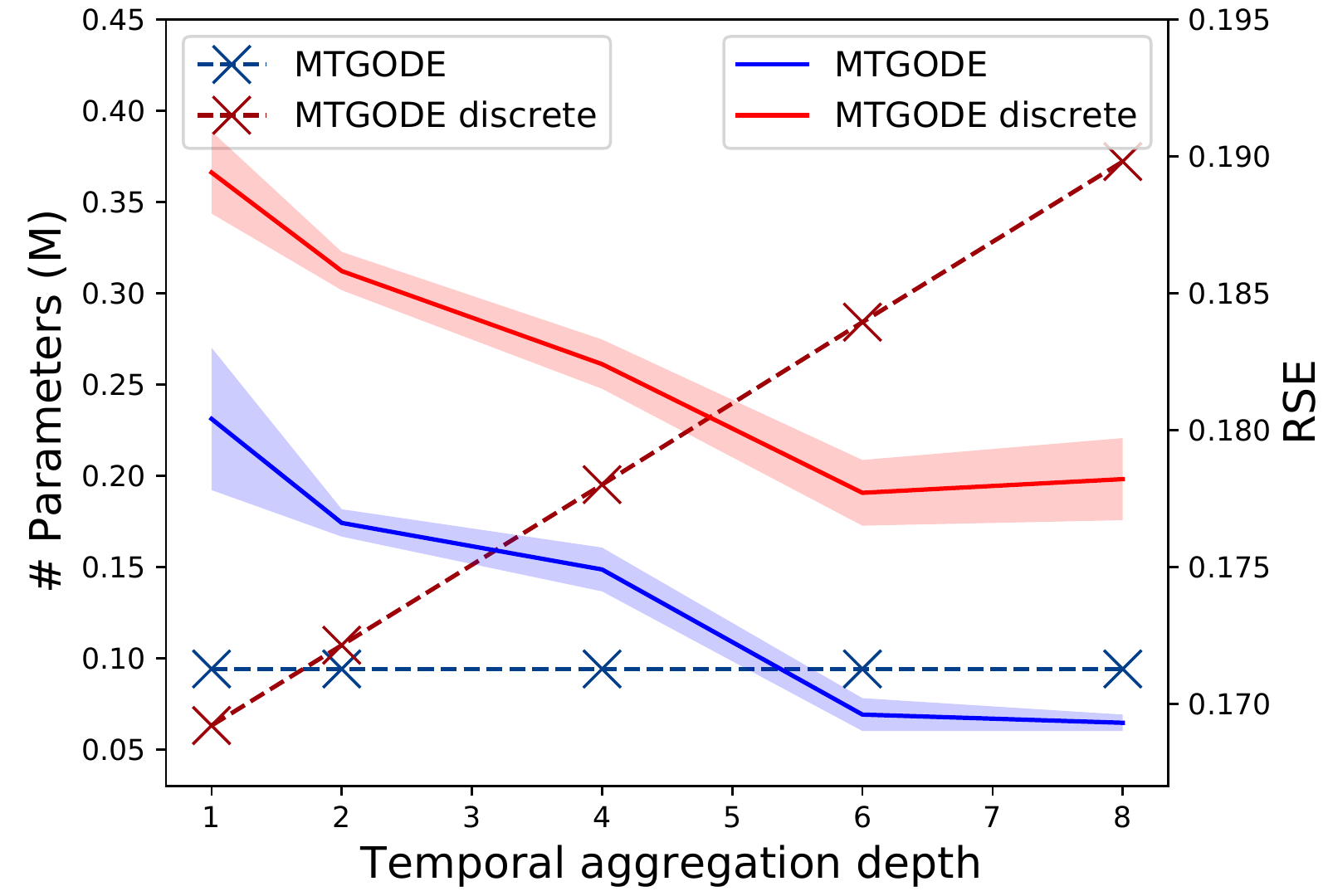} 
\label{subfig: temporal aggregation depth}
}
\subfigure{
\includegraphics[width=5cm]{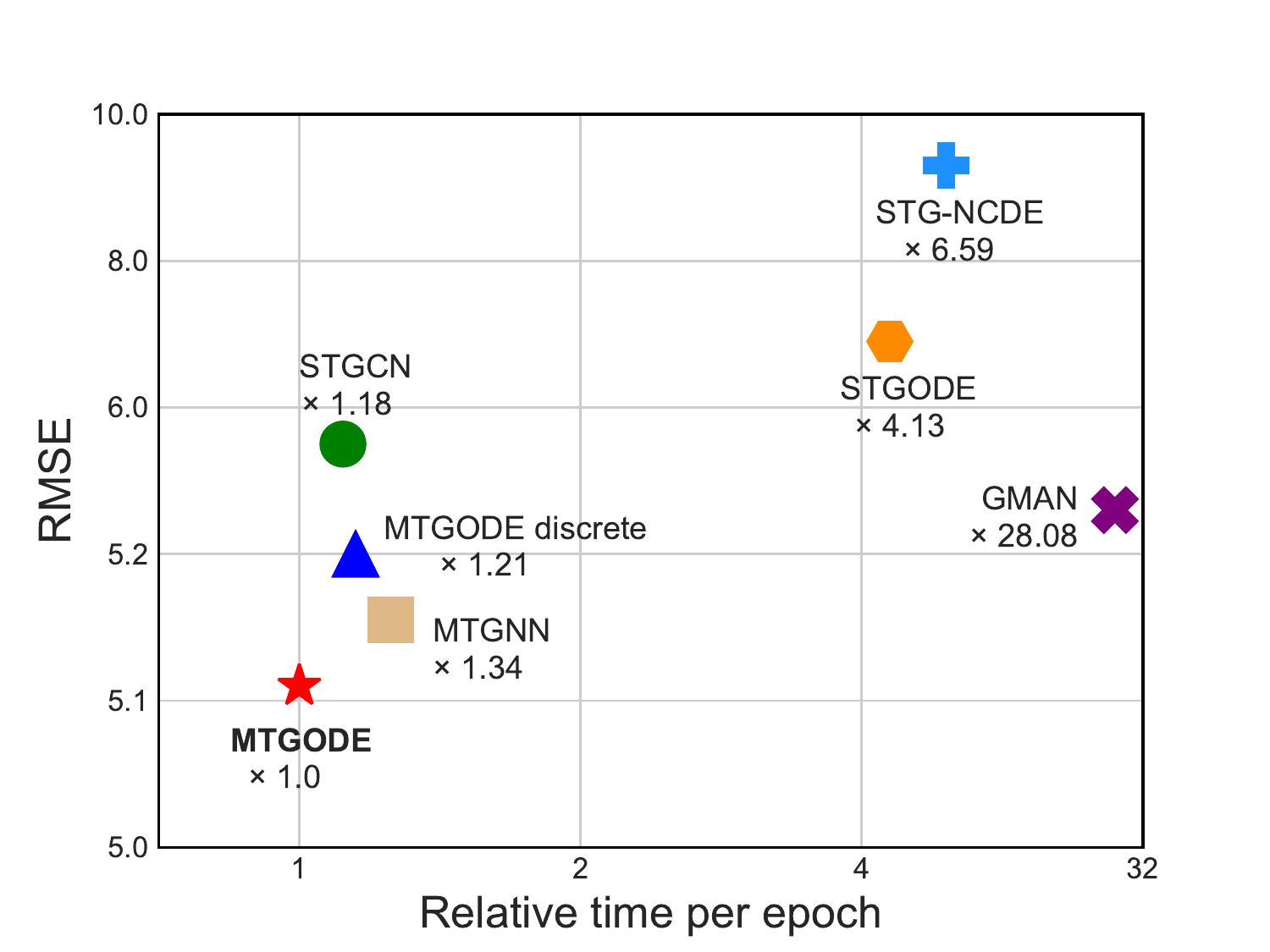}
\label{subfig: time per epoch}
}
\caption{
\textbf{Left and middle:} Model parameters and averaged performances w.r.t. graph propagation and temporal aggregation depths on Solar-Energy dataset. \texttt{MTGODE$^*$} is a variant of our method with the attentive transformation disabled in Equation \ref{eq: cgp attention}. \texttt{MTGODE discrete} denotes the discrete variant of our method by combining Equation \ref{eq: SGC} and the padding version of Equation \ref{eq: residual dilated tcn block}.
\textbf{Right:} \method vs. strong GNN-based discrete baselines on Metr-La dataset w.r.t. relative training time per epoch. The vertical axis shows the forecasting root mean square errors (horizon 3), and the \revision{horizontal} axis indicates the relative training time per epoch. \method surpasses dominant discrete methods by significant margins.
}
\label{fig:continuous regimes}
\end{figure*}

\noindent \textbf{Configuration.}
For \textit{multi-step forecasting}, we adopt Mean Absolute Error (MAE), Root Mean Square Error (RMSE), and Mean Absolute Percentage Error (MAPE) as our evaluation metrics \cite{wu2020connecting}. For \textit{single-step forecasting}, we follow \cite{lai2018modeling} and use Root Relative Squared Error (RSE) and Empirical Correlation Coefficient (CORR), where better performance is indicated by higher CORR and lower RSE values. All experiments are independently repeated ten times on Linux servers with two AMD EPYC 7742 CPUs and eight NVIDIA A100 GPUs. Averaged performances are reported.

\begin{itemize}
    \item \textbf{Single-step forecasting.} We choose the input length 168 and split all three benchmark time series datasets into training set (60\%), validation set (20\%), and testing set (20\%) chronologically. 
    The model is trained with Adam optimizer, batch size 4, and dropout rate 0.3. \revision{The hidden dimensions are fixed to 64.} 
    For Electricity and Traffic, our models are trained over 60 epochs with the base learning rate $10^{-3}$ and learning decays. Also, we let $\Delta t_{cta}=0.2$ and $\Delta t_{cgp}=0.5$.
    For Solar-Energy, we run 40 epochs with a fixed learning rate $10^{-4}$, and we have $\Delta t_{cta}=0.167$ and $\Delta t_{cgp}=0.25$.
    For graph learners, we adopt the settings suggested by \cite{wu2020connecting}.
    \vspace{1mm}
    \item \textbf{Multi-step forecasting.} We set the input and output lengths to 12 with the data split 70\%-10\%-20\%.
    On both datasets, we train 200 epochs using the Adam optimizer with a base learning rate of 0.001 and a dropout rate of 0.3.
    For Metr-La, the encoder and decoder hidden dimensions are 64 and 128. We use Euler solvers with integration time and \revision{step size} set to 1.0 and 0.25.
    For Pems-Bay, the hidden dimensions are 128. On this dataset, we use Runge-Kutta solvers with integration time and \revision{step size} set to 1.0 for simplicity. All experiments are with batch size 64 and learning rate decay. The configuration of graph learners is the same as in single-step forecasting.
\end{itemize}

\begin{table*}[t]
    \small
    \centering
 	\caption{The ablation study on three time series datasets. 
 	 We replace the CTA and CGP in \method with their discrete implementations, denoted as \texttt{w/o CTA} and \texttt{w/o CGP}. 
 	 For \texttt{w/o GSL}, the learned graph structure at each training step is replaced by a randomly generated adjacency matrix.
 	 We further remove the attentive and continuous regime in CGP to construct the \texttt{w/o CGP \& Attn}. A lower RSE and a higher CORR are expected.}
	\begin{tabular}{p{85 pt}<{} p{40 pt}<{\centering} p{40 pt}<{\centering} p{40 pt}<{\centering} p{40 pt}<{\centering} p{40 pt}<{\centering} p{40 pt}<{\centering}}
		
		\toprule
		    & \multicolumn{2}{c}{{Electricity}} & \multicolumn{2}{c}{{Traffic}} & \multicolumn{2}{c}{{Solar-Energy}} \\
          \cmidrule{2-7}
		  Method & RSE $\downarrow$ & CORR $\uparrow$ & RSE $\downarrow$ & CORR $\uparrow$ & RSE $\downarrow$ & CORR $\uparrow$ \\
		  
		  \midrule
		  
		  \method & \textbf{0.0727} & \textbf{0.9436} & \textbf{0.4088} & \textbf{0.9035} & \textbf{0.1686} & \textbf{0.9869} \\
		  
		  \midrule
		  \texttt{w/o GSL} & 0.0747 & 0.9414 & 0.4095 & 0.9011 & 0.1820 & 0.9847 \\
	      \texttt{w/o CTA} & 0.0777 & 0.9428 & 0.4126 & 0.9007 & 0.1893 & 0.9837 \\
	      \texttt{w/o CGP} & 0.0732 & 0.9427 & 0.4141 & 0.9015 & 0.1756 & 0.9857 \\
	      \texttt{w/o CGP \& Attn} & 0.0790 & 0.8979 & 0.4583 & 0.8820 & 0.1897 & 0.9831 \\
	
		\bottomrule
	
	\end{tabular}
	\label{table: ablation study}
\end{table*}

\subsection{Overall Comparisons} \label{subsec:overall comparisons}
We first report the results of different methods on different horizons for single-step forecasting in Table \ref{table: single-step forecasting results}. Specifically, we have two important observations:

\begin{itemize} 
    \item \revision{In general, \method achieves the best performance on three time series datasets, even when compared with HyDCNN and STG-NCDE, indicating its effectiveness in multivariate time series forecasting.}
	\item Our method significantly surpasses MTGNN in most cases with the same graph constructor, especially for long-term forecasting (i.e., horizons 6 and 12), demonstrating the superiority of our continuous regimes in capturing long-range and fine-grained spatial and temporal dependencies.
\end{itemize}

To further demonstrate the advantage of \method, we compare it with competitive GNN-based methods on two benchmark traffic datasets under the setting of multi-step forecasting, where all baselines use predefined graph structures only except for MTGNN, \revision{STG-NCDE,} and our method. We summarize the results in Table \ref{table: multi-step forecasting results}, from which we have the following observations: 

\begin{itemize} 
    \item \revision{Similar to single-step forecasting, our method consistently outperforms MTGNN and even STG-NCDE under this setting with a similar graph learning schema, which further confirms the effectiveness of \method in modeling multivariate time series data.}
    \item Our method demonstrates better performance compared with STGODE. This can be attributed to two reasons: Firstly, the proposed temporal ODE enables our approach to capture fine-grained temporal dynamics continuously. Secondly, our graph module is more expressive with the attentive transformation and free from graph priors; thus more robust to dataset biases.
    \item \method surpasses DCRNN, STGCN, and Graph WaveNet significantly without relying on graph priors. Compared to MRA-BCGN and GMAN, our method achieves the best or on-par performance, demonstrating its competitiveness.
\end{itemize}

\begin{figure*}[t]
\centering
\subfigure{
\includegraphics[width=.235\linewidth]{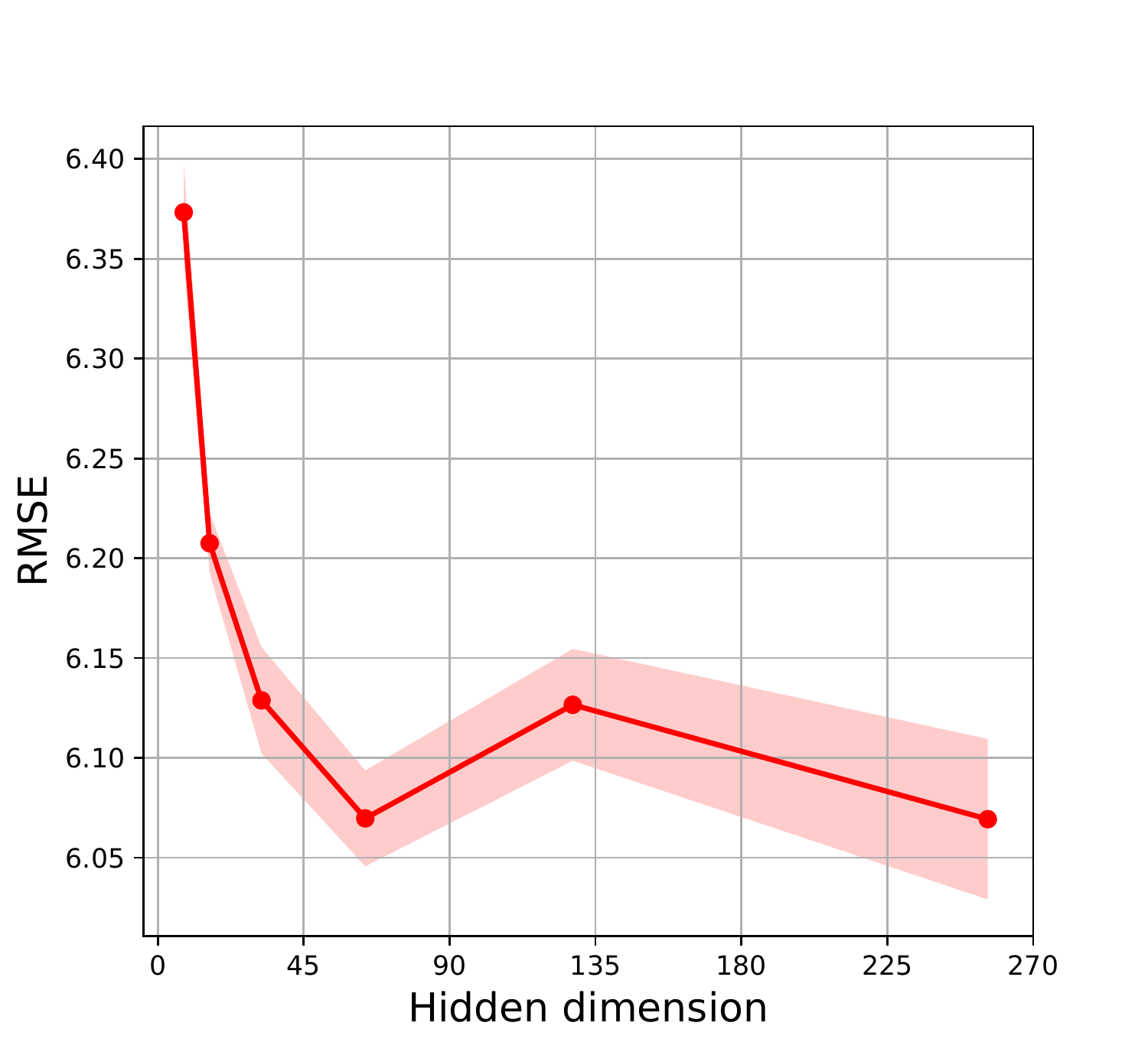}
\label{subfig: param_study_hidden}
}
\subfigure{
\includegraphics[width=.23\linewidth]{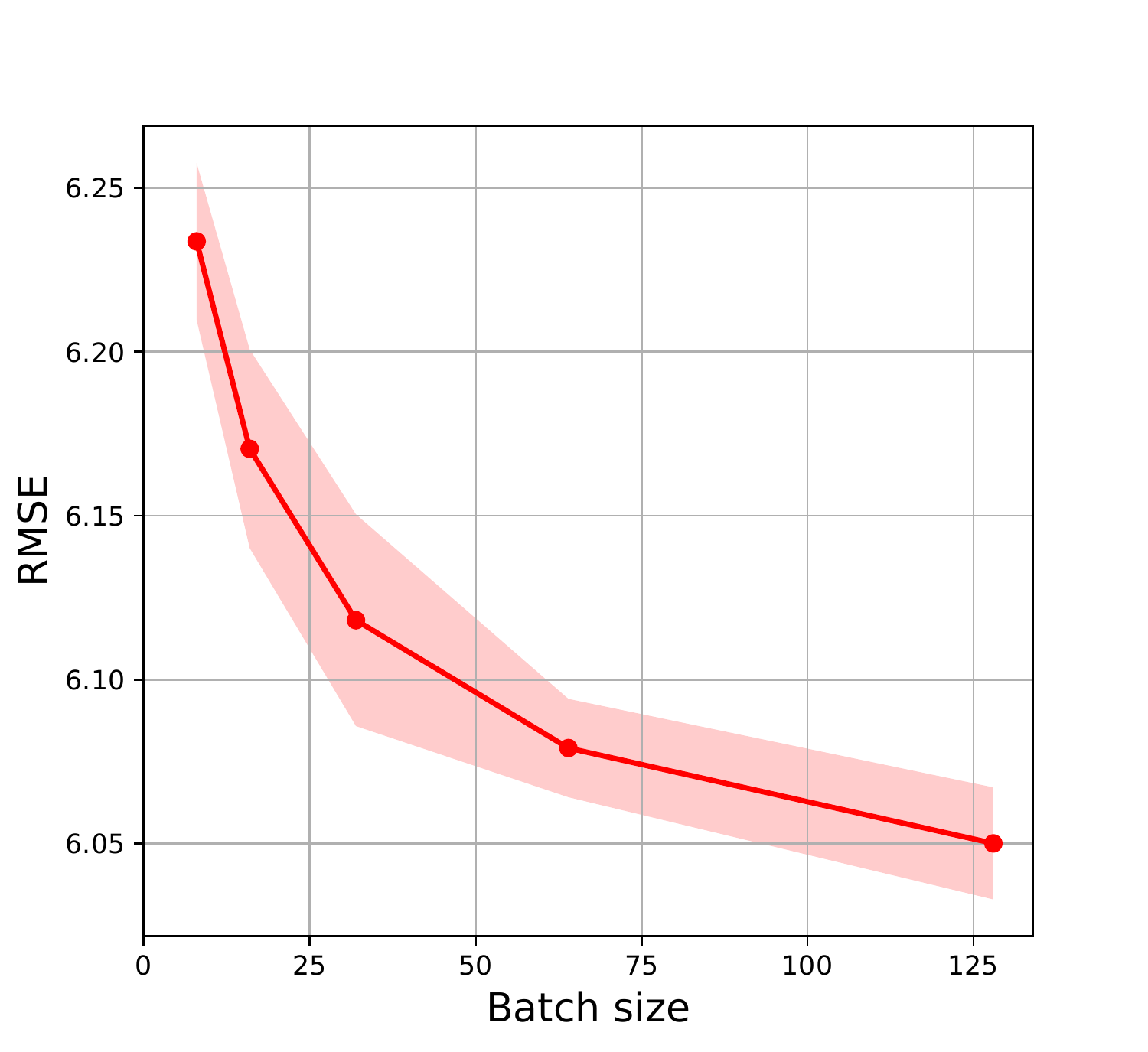}
\label{subfig: param_study_batch}
}
\subfigure{
\includegraphics[width=.23\linewidth]{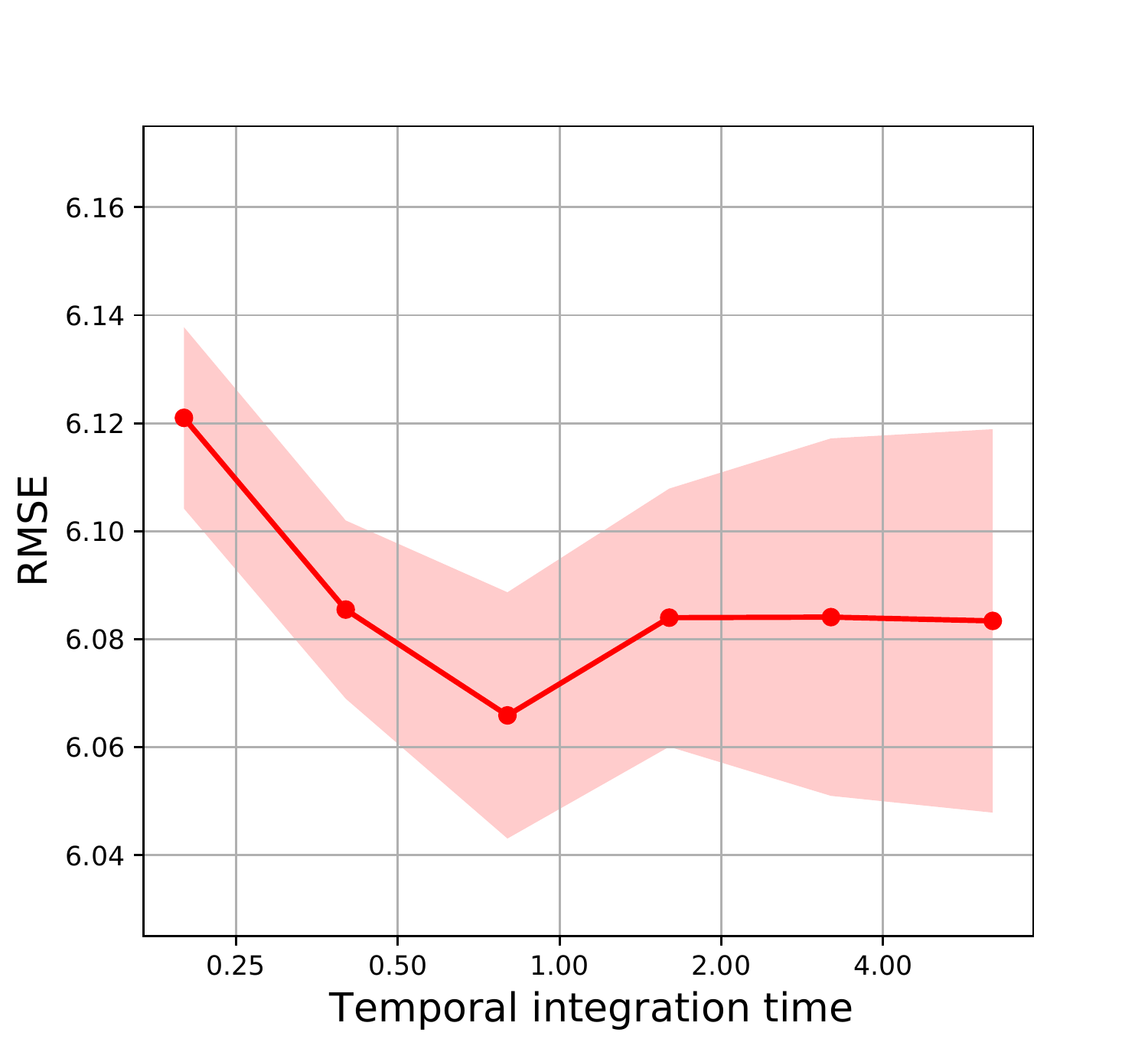}
\label{subfig: param_study_tcta}
}
\subfigure{
\includegraphics[width=.23\linewidth]{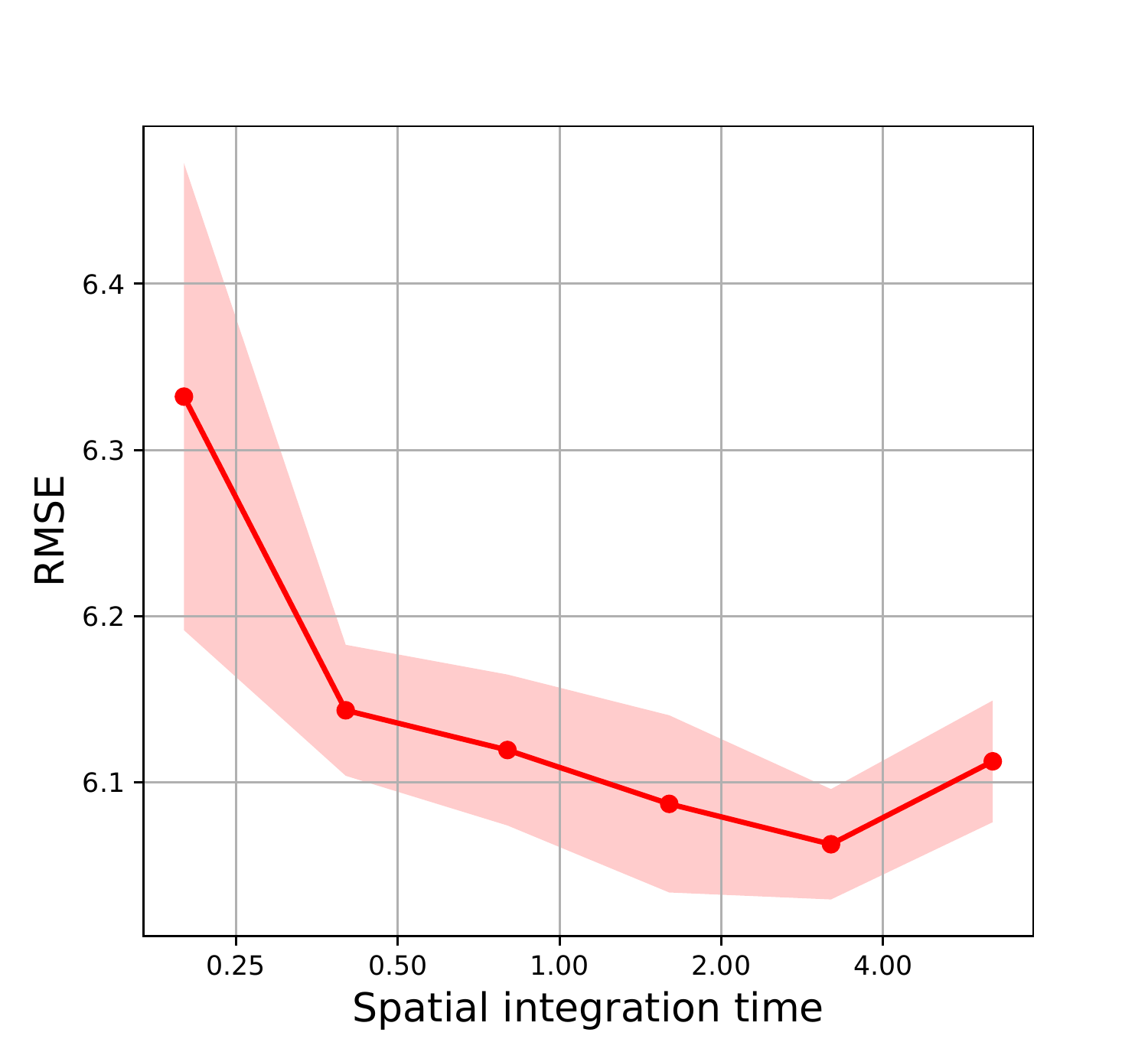}
\label{subfig: param_study_tcgp}
}
\caption{
Study on important parameters of \method on Metr-La dataset.
}
\label{fig:parameter study}
\end{figure*}

\subsection{Effectiveness of Two Continuous Regimes} \label{subsec:study on two ODEs}
To empirically validate the discussion in Subsection \ref{subsec:theoritical comparision} and study the behavioral differences between our method and its discrete variant, we dissect \method by comparing the model performance and the number of parameters with different graph propagation and temporal aggregation depths. 
Firstly, the left chart in Figure \ref{fig:continuous regimes} compares the proposed continuous graph propagation with its discrete implementation (Equation \ref{eq: SGC}). In particular, we disable the attentive transformation in this experiment to expose the essence of our proposed spatial ODE in Equation \ref{eq: cgp general}. Compared with \texttt{MTGODE discrete} (solid red curve), our method (solid blue line) is more robust (in terms of RSE) to the over-smoothing problem with increased propagation depths, where the gradually flattened performance curve and shrunk standard deviations indicate that our method allows the learned spatial representations to converge to a sweet spot by exploiting the long-range spatial dependencies, bringing significantly lower numerical errors (in terms of RSE) and better stability (w.r.t. standard deviations). 
At the same time, our method also demonstrates a better parameter efficiency (in terms of \# of parameters) compared with \texttt{MTGODE discrete}. It is worth noting that in this experiment, \texttt{MTGODE$^*$} and \texttt{MTGODE discrete} have constant parameters as feature propagation is parameterless.

The middle chart in Figure \ref{fig:continuous regimes} compares our method with its discrete variant by varying the temporal aggregation depth. We can observe that with the increase in aggregation depth, \method converges with decreased numerical errors (in terms of RSE). 
We can also observe that for \texttt{MTGODE discrete}, an increase in aggregation depth requires an increase in model parameters; hence it is complex.
In contrast, \method breaks this tie and thus allows a deeper aggregation to model more stable (w.r.t. standard deviations) and accurate temporal dynamics to capture fine-grained temporal patterns in a continuous and more parameter-efficient manner.
In Subsection \ref{subsec:computational efficiency}, we further demonstrate that \method is more computational and memory efficient than its discrete counterparts.

\subsection{Ablation Study}\label{subsec:ablation study}
We construct four variants of our method to study the effectiveness of core components. Specifically, \texttt{MTGODE w/o GSL} disables the dynamic graph construction mechanism, \texttt{MTGODE w/o CTA} and \texttt{MTGODE w/o CGP} replace the temporal and spatial ODEs with their discrete implementations to study the potency of two continuous regimes. 
\texttt{MTGODE w/o CGP \& Attn} further removes the attentive transformation in \texttt{MTGODE w/o CGP} to investigate the effectiveness of graph attentive transformation. In particular, the \texttt{MTGODE discrete} in Figure \ref{fig:continuous regimes} is equivalent to \texttt{MTGODE w/o CTA \& CGP \& Attn}, which has been investigated before \revision{so that} we omit this variant in ablation study.
The experimental results are in Table \ref{table: ablation study}, where our method equipped with all components has the best performance across all datasets. In particular, our spatial ODE with the attentive transformation benefits the model best for learning effective representations. 
Besides, the performance gains obtained by the CGP itself and the embedded dynamic graph construction mechanism are also notable.
A similar observation can also be made for CTA, where replacing it with its discrete version degrades the performance sharply.

\subsection{Parameters Sensitivity}\label{subsec:parameter}
Apart from the experiments on \revision{graph propagation and temporal aggregation depths} (\revision{i.e., the step sizes when solving \method since the integration time is fixed}) in Figure \ref{fig:continuous regimes}, we also conduct experiments on other important hyperparameters in \method, including temporal integration time $T_{cta}$, spatial integration time $T_{cgp}$, spatiotemporal encoder hidden dimension $D'$, and batch size $B$, to investigate their impacts on our model, as shown in Figure \ref{fig:parameter study}. Specifically, we have the following observations: (1) Moderately increasing the hidden state dimensions helps the model learning. We conjecture that this helps avoid the ODE trajectories intersecting with each other \cite{dupont2019augmented}, thus encouraging our model to learn smoother ODE functions that can be easier solved; (2) For a specific spatial or temporal propagation depth, we can find a sweet spot when selecting the spatial or temporal integration time. It may be because a short wall time hinders the convergence of the learned representations and a long time introduces relatively large numerical errors; (3) Within a reasonable range, e.g., from 32 to 128, moderately increasing the batch size improves the model performance. We hypothesize that a relatively large batch size in our method helps reduce the variances of mini-batch gradients, which reduces the impact of noise on the model training.

\subsection{Memory and Computational Efficiency}\label{subsec:computational efficiency}
\begin{figure}[b]
\centering
\includegraphics[width=3.9cm, height=3.8cm]{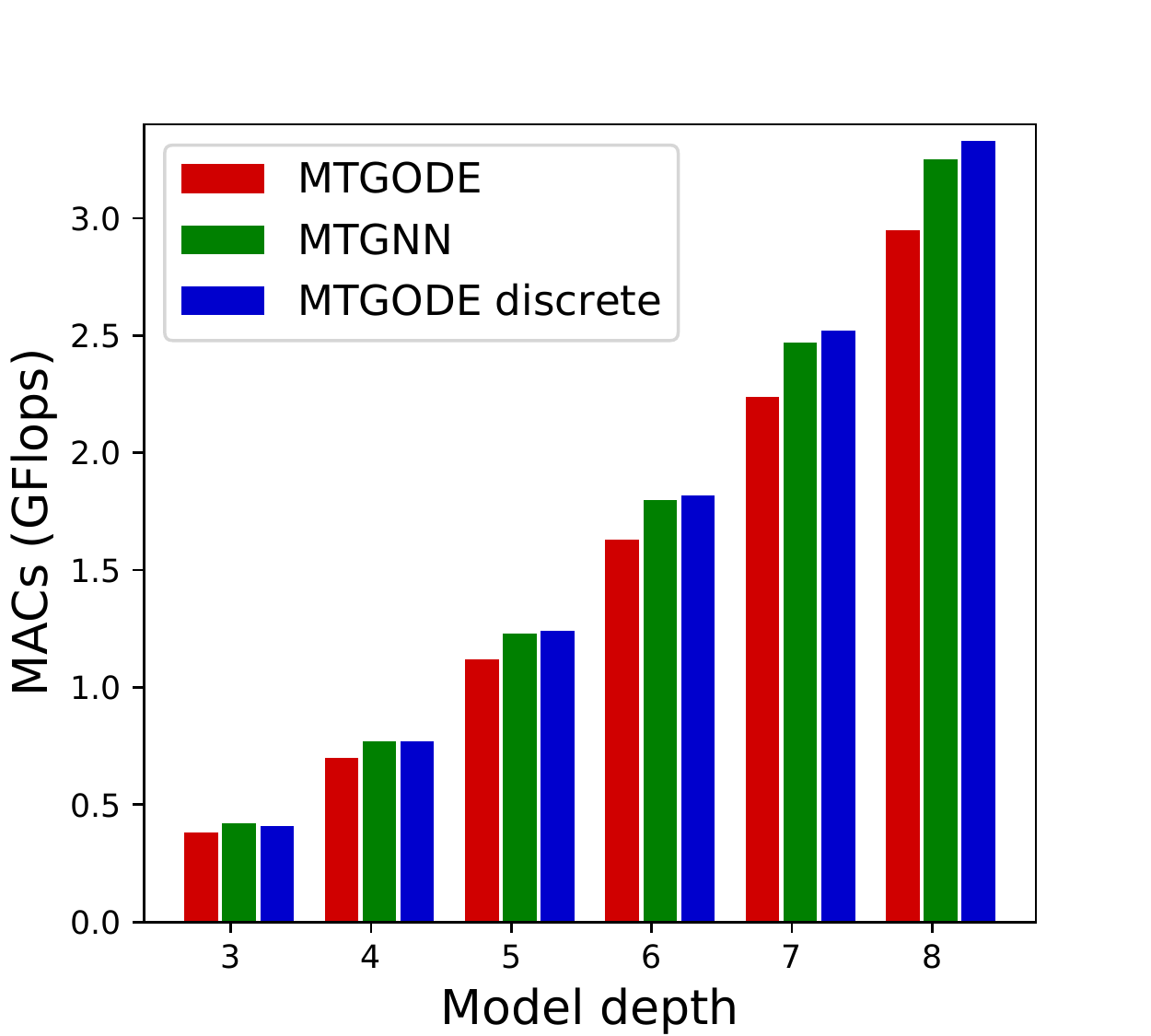}
\includegraphics[width=4cm, height=3.8cm]{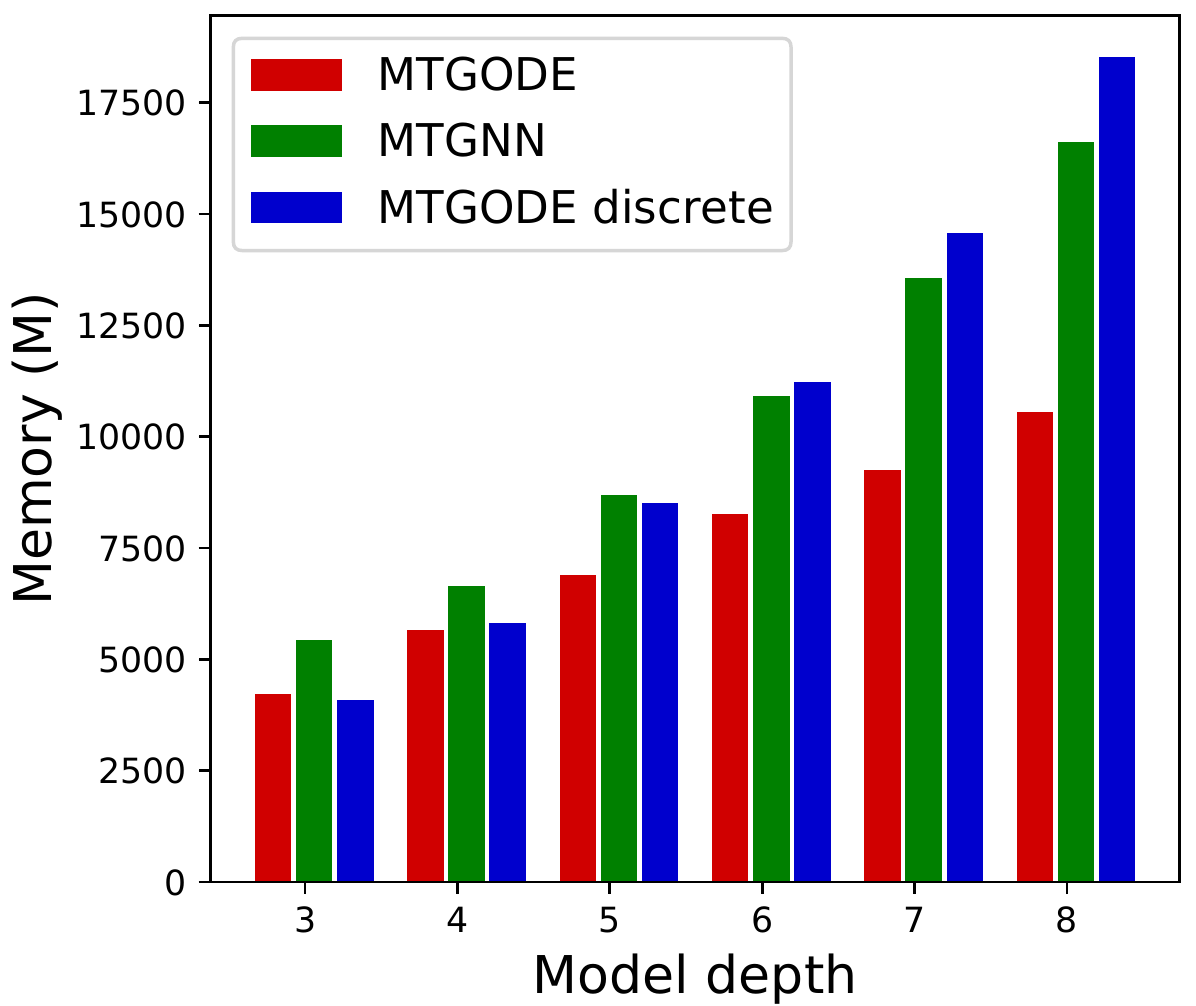}
\caption{The computational and memory overheads of \method and two discrete variants w.r.t. the model depth (i.e., temporal aggregation depth) on Metr-La dataset.}
\label{fig:macs and memory}
\end{figure}

In Figure \ref{fig:macs and memory}, we compare the required multiply-accumulate operations (MACs) and GPU memory of \method, its discrete variant, and MTGNN \cite{wu2020connecting}. In general, our method constantly has lower computational and memory overheads than \texttt{MTGODE discrete} and MTGNN, especially for larger model depths, demonstrating the computational and memory efficiency of \method. In comparison, discrete methods, e.g., MTGNN and our discrete variant, have more complex and discrete neural architectures, which inevitably introduce more intermediate operations and trainable parameters, resulting in higher computational and memory costs.
In particular, we find that MTGNN is slightly more efficient than \texttt{MTGODE discrete} in terms of large model depth because the latter one adopts the padding version of Equation \ref{eq: residual dilated tcn block}, which inevitably involves more parameters in the following $L-1$ temporal convolution layers except for the first layer. Although \method is also based on this padding trick, it is still more computationally and memory efficient than MTGNN.
\revision{In the rightmost chart in Figure \ref{fig:continuous regimes}, we further demonstrate that the proposed method is more efficient than discrete methods to provide better forecasting results. Specifically, in this experiment, we slightly increase the spatial and temporal step size in our method (i.e., $0.25 \rightarrow 0.34$ and $0.25 \rightarrow 0.5$ in CTA and CGP processes, respectively) to trade model precision for speed, \method remains surpassing strong GNN-based baselines by a significant margin in terms of forecasting errors.}

\section{Conclusion} \label{sec:conclusion}
\revision{Given the shortcomings of prior arts in multivariate time series forecasting, we investigate using neural ordinary differential equations and dynamic graph structure learning to model the continuous latent spatial-temporal dynamics of arbitrary multivariate time series. By solving the intersecting continuous graph propagation and temporal aggregation processes, our method allows the model to learn more expressive representations efficiently without relying on graph priors, showing better potential in real-world applications.}
Apart from the empirical justifications, we also theoretically analyze the main properties of our method and further demonstrate that it is more effective and efficient than the existing discrete approaches.


\bibliographystyle{IEEEtran}
\bibliography{IEEEabrv,main}

\appendices

\section{Proof of Property 1}\label{appx: property 1}
Given a simplified graph feature propagation in Equation \ref{eq: SGC} in the main text, we give proof that it is characterized by the following ODE.
\begin{equation}
\begin{aligned}
\frac{\displaystyle\mathrm{d}\mathbf{H}^G(t)}{\displaystyle\mathrm{d}t} &= (\widehat{\mathbf{A}}-\mathbf{I}_N) \ \mathbf{H}^G(t), \\
&= -\mathbf{L}\mathbf{H}^G(t), \ t \in \mathbb{R}_{0}^{+},
\end{aligned}
\end{equation}
where $\mathbf{L}=\mathbf{I}_N - \widehat{\mathbf{A}}$ denotes the normalized graph Laplacian. Regarding the above ODE, it can be naturally viewed as a general graph heat diffusion process with the Laplacian $\mathbf{L}$ \cite{wang2021dissecting}, where the closed-form solution is:
\begin{equation}
\frac{\displaystyle\mathrm{d}\mathbf{H}^G(t)}{\displaystyle\mathrm{d}t} = -\mathbf{L}\mathbf{H}^G(t)
\ \ \ \ \Rightarrow \ \ \ \ 
\mathbf{H}^G(t) = e^{-t\mathbf{L}}\mathbf{H}^G(0).
\label{eq: heat kernel solution}
\end{equation}

In the above equation, $e^{-t\mathbf{L}}$ is known as the heat kernel. For the graph Laplacian $\mathbf{L}=\mathbf{I}_N - \widehat{\mathbf{A}}$, if $\widehat{\mathbf{A}}$ is symmetrically normalized, we have $\mathbf{L}=\mathbf{I}_N - \widetilde{\mathbf{D}}^{-\frac{1}{2}} \widetilde{\mathbf{A}} \widetilde{\mathbf{D}}^{-\frac{1}{2}} = \widetilde{\mathbf{D}}^{-\frac{1}{2}} (\widetilde{\mathbf{D}} - \widetilde{\mathbf{A}}) \widetilde{\mathbf{D}}^{-\frac{1}{2}}$, which is symmetric and positive semi-definite. Thus, the eigendecomposition of $\mathbf{L}$ can be defined as follows:
\begin{equation}
\mathbf{L} = \mathbf{U}\mathbf{\Lambda}\mathbf{U}^\top,
\end{equation}
where $\mathbf{U}$ is an orthogonal matrix of eigenvectors, and $\mathbf{\Lambda}$ is a diagonal matrix that consists of eigenvalues $\lambda_i \geq 0$. Based on this, the heat kernel can be decomposed as follows based on the Taylor expansion:
\begin{equation}
\begin{aligned}
e^{-t\mathbf{L}} &= \sum_{k=0}^{\infty} \frac{1}{k!}(-t\mathbf{L})^k
= \sum_{k=0}^{\infty} \frac{t^k}{k!}\big[\mathbf{U}(-\mathbf{\Lambda})\mathbf{U}^\top \big]^k \\
&= \mathbf{U}\big[\sum_{k=0}^{\infty} \frac{t^k}{k!} (-\mathbf{\Lambda})^k\big]\mathbf{U}^\top
= \mathbf{U}e^{-t\mathbf{\Lambda}}\mathbf{U}^\top.
\end{aligned}
\end{equation}

Thus, the eigendecomposition of the heat kernel can be easily obtained:
\begin{equation}
   e^{-t\mathbf{L}} = \mathbf{U} 
   \begin{pmatrix} 
   e^{-t\lambda_1} & 0 & \cdots & 0 \\
   0 & e^{-t\lambda_2} & \cdots & 0 \\
   \vdots & \vdots & \ddots &  \vdots \\
   0 & 0 & \cdots & e^{-t\lambda_N}
   \end{pmatrix} 
   \mathbf{U}^\top,
\end{equation}
where for each of eigenvalues $e^{-t\lambda_i}$, they satisfy the following property when $t \rightarrow \infty$:
\begin{equation}
\lim_{t\to\infty} e^{-t\lambda_i} = 
\left\{
\begin{aligned}
& 0, \ \ \ \text{if} \ \ \lambda_i > 0 \\
& 1, \ \ \ \text{if} \ \ \lambda_i = 0.
\end{aligned}
\right.
\end{equation}

As such, given a graph that $\mathbf{A} \neq \mathbf{I}_N$, increasing the propagation depth in Equation \ref{eq: SGC} in the main text will inevitably lead to the over-smoothing problem, where the eigenvalues are zeroed with $K=T \rightarrow \infty$:
\begin{equation}
\lim_{T\to\infty} \mathbf{H}^G(t) = e^{-T\mathbf{L}} \mathbf{H}^G(0) = \mathbf{0}.
\end{equation}

Conversely, in the proposed spatial ODE, we disengage the coupling between the propagation depth $K$ and terminal (integration) time $T_{cgp}$ by making $K=T_{cgp}/\Delta t_{cgp}$. Thus, it is possible to increase the propagation depth without letting  $T_{cgp} \rightarrow \infty$, which ensures the convergence of the learned spatial representations.

\section{Proof of Property 2}\label{appx: property 2}
Similar to the proof in Appendix \ref{appx: property 1}, given a terminal (integration) time $T_{cgp}$, the proposed spatial ODE can be viewed as a general graph heat diffusion process with the Laplacian $\mathbf{L}$, where the closed-form solution is given by:
\begin{equation}
\mathbf{H}^G(T_{cgp}) = e^{-T_{cgp}\mathbf{L}}\ \mathbf{H}^G(0).
\label{eq: heat kernel solution 2}
\end{equation}

For the heat kernel $e^{-T_{cgp}\mathbf{L}}$, its can be expanded in a Taylor series:
\begin{equation}
e^{-T_{cgp}\mathbf{L}} = \sum_{k=0}^{\infty} \frac{T_{cgp}^k}{k!}(-\mathbf{L})^k.
\end{equation}

Accordingly, Equation \ref{eq: heat kernel solution 2} can be reformulated as follows:
\begin{equation}
\mathbf{H}^G(T_{cgp}) = \Big[\sum_{k=0}^{\infty} \frac{T_{cgp}^k}{k!}(-\mathbf{L})^k\Big] \ \mathbf{H}^G(0).
\label{eq: heat kernel solution 3}
\end{equation}

Considering an Euler solver is applied, the numerical solution of the above equation after $K$ propagation steps is:
\begin{equation}
\widehat{\mathbf{H}}^G(T_{cgp}) = (\mathbf{I}_N - \frac{T_{cgp}}{K}\mathbf{L})^k \ \mathbf{H}^G(0).
\label{eq: solved numerical solution in appendix}
\end{equation}

Thus, the numerical errors between the analytical and solved numerical solutions (i.e., Equation \ref{eq: heat kernel solution 3} and \ref{eq: solved numerical solution in appendix}) can be simply defined as follows:
\begin{equation}
\mathbf{E}^{(K)}_{T_{cgp}} = \mathbf{H}^G(T_{cgp}) - \widehat{\mathbf{H}}^G(T_{cgp}).
\label{eq: numerical errors}
\end{equation}

According to \cite{wang2021dissecting}, we have $\mathbf{E}^{(K)}_{T_{cgp}}$ to be upper bounded by the following inequation:
\begin{equation}
\left \| \mathbf{E}^{(K)}_{T_{cgp}} \right\| = \frac{T_{cgp} \left \| \mathbf{L} \right\| \left \| \mathbf{H}^G(0) \right\|}{2K}(e^{T_{cgp} \left \| \mathbf{L} \right\|} - 1)
\end{equation}

Thus, for a fixed terminal time $T_{cgp}$, we can easily find that $\mathbf{E}^{(K)}_{T_{cgp}} \rightarrow 0$ by letting the propagation depth $K \rightarrow \infty$. Conversely, let $T=K \rightarrow \infty$ in Equation \ref{eq: SGC} in the main text will lead $\mathbf{E}^{(K)}_{T_{cgp}} \rightarrow \infty$.

\end{document}